\PassOptionsToPackage{square,comma,numbers,sort&compress}{natbib}
\documentclass[10pt,twocolumn,letterpaper]{article}

\usepackage[pagenumbers]{cvpr} 

\usepackage[dvipsnames]{xcolor}

\definecolor{cvprblue}{rgb}{0.21,0.49,0.74}
\usepackage[pagebackref,breaklinks,colorlinks,citecolor=cvprblue]{hyperref}

\usepackage{multirow}
\usepackage{graphicx}
\usepackage[dvipsnames]{xcolor}

\newcommand\issuefinder[0]{{\color{BrickRed}Issue Finder}}
\newcommand\datafeeder[0]{{\color{Green}Data Feeder}}
\newcommand\modelupdater[0]{{\color{Orange}Model Updater}}
\newcommand\verification[0]{{\color{Purple}Verification}}

\title{AIDE: An Automatic Data Engine for Object Detection in Autonomous Driving}

\author{Mingfu Liang$^1$ \quad Jong-Chyi Su$^2$ \quad Samuel Schulter$^2$ \quad Sparsh Garg$^2$ \quad Shiyu Zhao$^3$ \\
\quad Ying Wu$^1$ \quad Manmohan Chandraker$^{2,4}$ \\
$^1$ Northwestern University \quad $^2$ NEC Laboratories America \quad $^3$ Rutgers University \quad $^4$ UC San Diego
}

\begin{document}
\maketitle
\begin{abstract}

Autonomous vehicle (AV) systems rely on robust perception models as a cornerstone of safety assurance. However, objects encountered on the road exhibit a long-tailed distribution, with rare or unseen categories posing challenges to a deployed perception model. This necessitates an expensive process of continuously curating and annotating data with significant human effort. We propose to leverage recent advances in vision-language and large language models to design an Automatic Data Engine (AIDE) that automatically identifies issues, efficiently curates data, improves the model through auto-labeling, and verifies the model through generation of diverse scenarios. This process operates iteratively, allowing for continuous self-improvement of the model. We further establish a benchmark for open-world detection on AV datasets to comprehensively evaluate various learning paradigms, demonstrating our method's superior performance at a reduced cost.{\let\thefootnote\relax\footnote{{This work was part of Mingfu's internship at NEC Labs America. Email: \href{mingfuliang@u.northwestern.edu}{mingfuliang2020@u.northwestern.edu}}}}
\end{abstract}

\section{Introduction}
Autonomous vehicles~(AVs) operate in an ever-changing world, encountering diverse objects and scenarios in a long-tailed distribution. 
This open-world nature poses a significant challenge for AV systems since it is a safety-critical application where reliable and well-trained models must be deployed.
The need for continuous model improvement becomes apparent as the environment evolves, demanding adaptability to handle unexpected events.
Despite the wealth of data collected on the road every minute, its effective utilization remains low due to challenges in discerning which data to leverage. 
While solutions exist for this in industry~\cite{Tesla, Cruise}, they are often trade secrets and presumably require significant human effort.
Hence, developing a comprehensive automated data engine can lower entry barriers for the AV industry.

\begin{figure}[th!]
  \centering
  \includegraphics[width=0.48\textwidth]{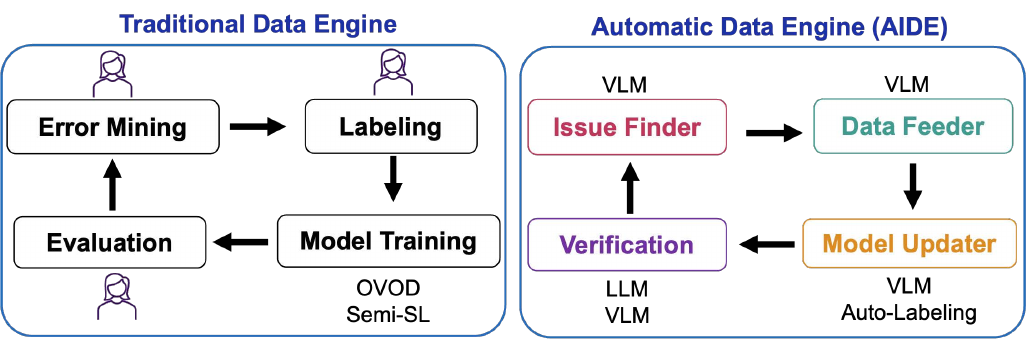}
  \includegraphics[width=0.4\textwidth]{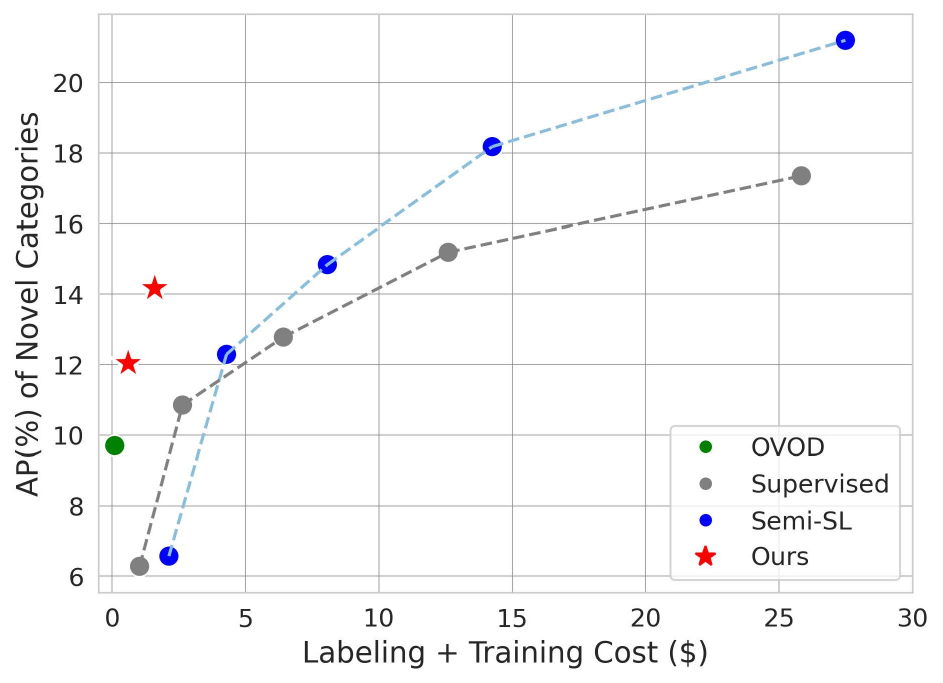}
  \caption{\textbf{Top:} Components for DevOp systems for autonomous driving. \textbf{Bottom:} With our automatic data system, we can achieve similar performance with less labeling and training costs.}
  \label{fig:teaser}
  \vspace{-10pt}
\end{figure}

Designing automated data engines can be challenging, but the existence of Vision-Language Models~(VLMs) and Large Language Models~(LLMs) allows new avenues to these hard problems. 
A traditional data engine can be broken down into finding issues, curating and labeling data, model training, and evaluation, all of which can benefit from automation.
In this paper, we propose an Automatically Improving Data Engine (called AIDE) that leverages VLMs and LLMs to automate the data engine. 
Specifically, we use VLMs to identify the issue, query relevant data, auto-label data, and verify together with LLMs.
The high-level steps are shown in Fig.~\ref{fig:teaser} top. 

In contrast to traditional data engines that rely heavily on extensive human labeling and intervention, AIDE automated the process by utilizing pre-trained VLMs and LLMs. Different from other confidential solutions in industry~\cite{Tesla,Cruise}, we provide our efficient solutions to lower the entry barrier.
While open-vocabulary object detection~(OVOD) methods~\cite{labonte2023scaling,minderer2022simple} do not require any human annotations, they are a good starting point for detecting novel objects but their performances fall short on AV datasets compared to supervised methods. 
Another line of research on minimizing labeling costs is semi-supervised learning~\cite{liu2021unbiased,liu2022unbiased} and active learning~\cite{sener2017active,elezi2022not,kothawade2022talisman,lyu2023box}.
Although they generate pseudo-labels, the vast amount of data collected on the road is still not fully utilized, in contrast with our method which leverages pre-trained VLMs and LLMs for better data utilization.

The detailed steps of AIDE are shown in Fig.~\ref{fig:overview}. 
In the \issuefinder, we use a dense captioning model to describe the image in detail, then match if the objects in the description are included in the label spaces or the predictions. 
This is based on the reasonable but previously unexploited assumption that large image captioning models are more robust starting points in zero-shot settings than OVOD (Tab.~\ref{tab:ab_issuefinder}).
The next step is to find relevant images that could contain the novel category using our \datafeeder. 
We find that VLM gives more accurate image retrieval than using image similarity to retrieve images~(Tab.~\ref{tab:ab_datafeeder}).
We then use our existing label space plus the novel category to prompt the OVOD method, i.e., OWL-v2~\cite{minderer2023scaling}, to generate predictions on the queried images. 
To filter these pseudo predictions, we use CLIP to perform zero-shot classification on the pseudo-boxes to generate pseudo-labels for the novel categories. 
Last, we exploit the LLM, e.g., ChatGPT~\cite{ChatGPT}, in \verification{} to generate diverse scene descriptions given the novel objects. 
Given the generated description, we again use VLM to query relevant images to evaluate the updated model. 
To ensure the correctness, we ask humans to review if the predictions of the novel categories are correct. If it is not, we ask humans to provide ground-truth labels, which are used to further improve the model.~(Fig.~\ref{fig:verification})

To verify the effectiveness of our AIDE, we propose a new benchmark on existing AV datasets to comprehensively compare our AIDE with other paradigms. With our \issuefinder, \datafeeder{}, and \modelupdater{}, we bring 2.3\% Average Precision~(AP) improvement on the novel categories compared with OWL-v2 without any human annotations and also surpass OWL-v2 by 8.9\% AP on known categories (Tab.~\ref{tab:main_table_avg_only}).
We also show that with a single round of \verification{}, our automatic data engine can further bring 2.2\% AP on novel categories without forgetting the known categories, as shown in Fig.~\ref{fig:teaser}.
To summarize, our contributions are two-fold: 

\begin{figure*}[ht!]
  \centering
  \includegraphics[width=1.0\linewidth]{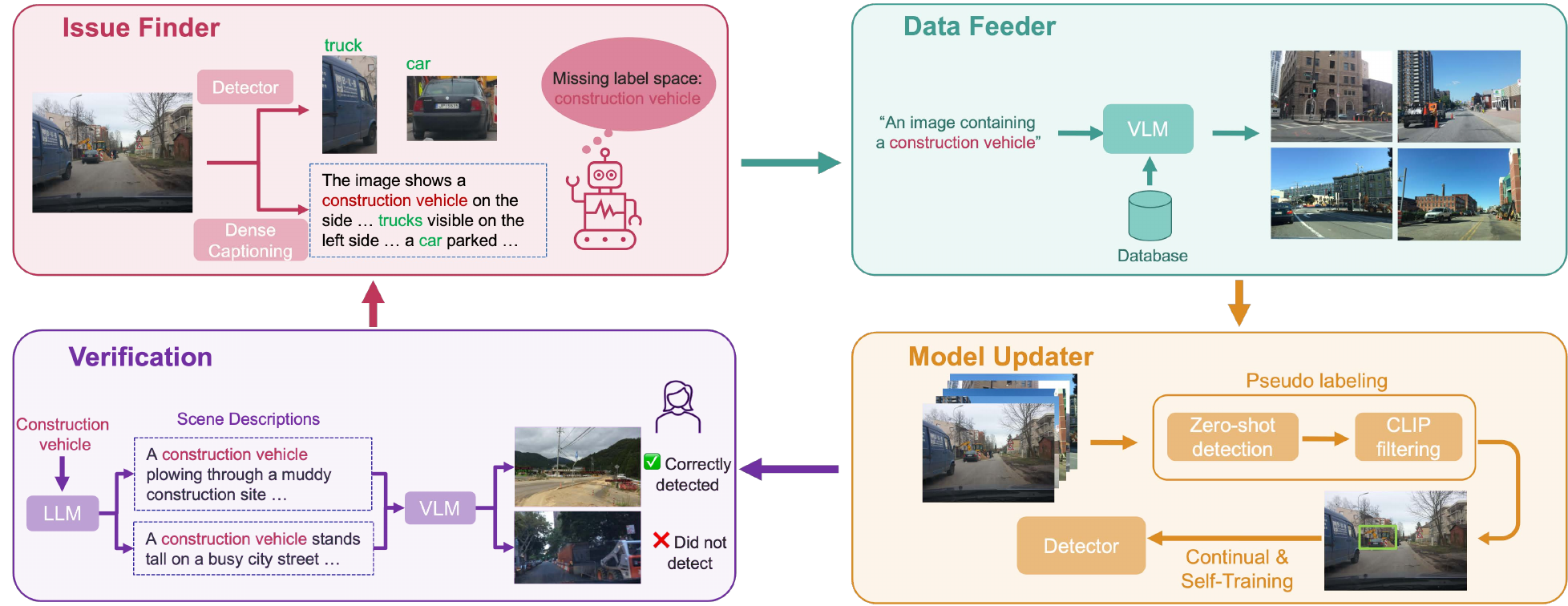}
  \caption{Our design of the automatic data engine includes \issuefinder, \datafeeder, \modelupdater, and \verification. The \issuefinder{} automatically identifies novel categories using the dense captioning model. In the \datafeeder{}, we employ VLMs to efficiently search for relevant data for training, significantly reducing the inference time for generating pseudo-labels in the subsequent steps and filtering out unrelated images for training. The model is updated in the \modelupdater{} using auto-labeling by VLMs, enabling the recognition of novel categories without incurring any labeling costs. To verify the model, in \verification{}, we use LLMs to generate descriptions of variations in scenarios and then assess predictions on images queried by VLMs.
  }
  \label{fig:overview}
\end{figure*}

\begin{itemize}
    \item  We propose a novel design paradigm for an automatic data engine for autonomous driving as automatic data querying and labeling with VLM and continual learning with pseudo labels. When scaling up for novel categories, this approach achieves an excellent trade-off between detection performance and data cost. 
    \item We introduce a new benchmark to evaluate such automated data engines for AV perception that allows combined insights across multiple paradigms of open vocabulary detection, semi-supervised, and continual learning.
\end{itemize}

\section{Related Works}
\textbf{Data Engine for Autonomous Vehicles~(AV)} Exploiting large-scale data collected by AV is crucial to speed up the iterative development of the AV system~\cite{chen2023end}. 
Existing literature mostly focuses on developing general~\cite{chen2013neil, mitchell2018never} learning engines or specific~\cite{kirillov2023segment} data engines, and most of them~\cite{yang2021auto4d, qi2021offboard} mainly focus on the model training part. 
However, a fully functional AV data engine requires issue identification, data curation, model retraining, verification, etc. 
A thorough examination reveals a lack of systematic research papers or literature that delves deeply into AV data engines in academia, where a recent survey~\cite{chen2023end} also underscores the lack of study in this context. 
On the other hand, existing solutions~\cite{Tesla, Cruise} for AV data systems mainly rely on the design of data infrastructure and still need lots amount of human effort and intervention, thus limiting their maintenance simplicity, affordability, and scalability. 
In contrast, the present paper exploits the burgeoning progress of vision language models~(VLMs)~\cite{thengane2022clip,li2023otter,li2023blip} to design our data engine, where their strong open-world perception capability largely improves our engine's extendability and makes it more affordable to scale up our AVs on detecting novel categories. 
To our best knowledge, this paper is also the first work that provides a systematic design of data engines for AVs with the integration of VLMs.

\noindent{\textbf{Novel Object Detection}} Conventional 2D object detection has made enormous progress~\cite{girshick2015fast,carion2020end} in the last decades, while its closed-set label space makes unseen category detection infeasible.  On the other hand, open-vocabulary object detection~(OVOD)~\cite{perez2020incremental,dhamija2020overlooked,kim2022learning,saito2022learning,bravo2022localized,cai2022x,lin2023learning,zheng2022towards,joseph2021towards,bao2022discovering,vaze2022gcd,fini2021unified,gu2022openvocabulary,kim2023region,kuo2023openvocabulary,zhang2023simple,minderer2022simple} methods promises to detect anything by a simple text prompting. However, their performances are still inferior to closed-set object detection since they must balance the specificity of pre-trained categories and the generalizability of unseen categories. To scale up the capacity of open-vocabulary detector~(OVD), recent works either pre-train OVD with weak annotations~(e.g., image captions)~\cite{li2022grounded}, or perform self-training on daily object datasets~\cite{zhao2022exploiting,gupta2019lvis} or web-scale datasets~\cite{minderer2022simple,chen2023pali}. However, balancing the trade-off between improving the novel categories while mitigating the catastrophic forgetting of the known categories is still an open problem that has not been resolved~\cite{minderer2023scaling}, making it hard to adapt to task-specific applications like autonomous driving.

On the other hand, limited research has focused on novel object detection for AVs. This is especially crucial because a false-negative detection of unseen objects may result in fatal consequences for AVs.
Existing OVOD methods mostly benchmark on datasets of general objects~\cite{lin2014microsoft,gupta2019lvis} while putting little attention on AV datasets~\cite{nuscenes2019, Sun_2020_CVPR, Cordts2016Cityscapes, Geiger2012CVPR, bdd100k, neuhold2017mapillary}. Different from the pursuit of generality in OVOD, perception in AVs has its domain concerns oriented from the image-capturing process by on-car cameras and the object categories due to the scene prior (e.g., road/street objects), which demands task-specific design to enable efficient and scalable system to iteratively enhance AVs on detecting novel objects during its lifecycle. 
To strike a better trade-off between specificity and generality, our proposed AIDE iteratively extends the closed-set detector's label space so that we can retain decent performance on both novel and known categories for better detection.

\noindent{\textbf{Semi-Supervised Learning (Semi-SL) and Active Learning (AL)}} As AVs keep collecting data in operation, a native solution to enable novel category detection is to manually identify the novel category over a collected unlabeled data pool, label them, and then train the detector. Semi-SL~\cite{gao2020consistency,liu2021unbiased,liu2022unbiased,kothawade2022talisman,zhang2023semi,wang2023consistent,liu2022open} and AL~\cite{ren2021survey,segal2022just,elezi2022not,qi2021offboard,jiang2022improving,ning2022active,lyu2023box} seem to help as they require only a small amount of labeled data to initialize the training. 
However, labeling even a small amount of data for novel categories will be challenging and costly when given a vast amount of unlabeled data~\cite{zhu2003combining,segal2022just,elezi2022not,sadat2021diverse,liu2023mixteacher} by AVs. 
Moreover, both Semi-SL and AL assume that the labeled and unlabeled data come from the same distribution~\cite{mi2022active,borsos2021semi,gao2020consistency} and share the same label space.
However, this assumption does not hold when new categories emerge, inevitably leading to changes in the label space.
Naive fine-tuning of the detector only on the novel categories will lead to catastrophic forgetting~\cite{shmelkov2017incremental,wang2021wanderlust,feng2022overcoming} of known categories learned previously. 
However, Semi-SL methods for object detection do not consider continual learning, while existing continual semi-supervised learning methods~\cite{wang2021ordisco,smith2021memory,boschini2022continual,kang2023soft} are also specific to image classification, which is not applicable for object detection. 

\section{Method}
This section demonstrates our proposed AIDE, composed of four components: \issuefinder, \datafeeder, \modelupdater, and \verification. The \issuefinder{} automatically identifies missing categories in the existing label space by comparing detection results and dense captions given an image. This triggers the \datafeeder{} to perform text-guided retrieval for relevant images from the large-scale image pool collected by AVs. The \modelupdater{} then automatically labels queried images and continuously trains the novel category with pseudo-labels on the existing detector. The updated detector is then passed to the \verification{} module to evaluate under different scenarios and trigger a new iteration if needed. We outline our systematic design in Fig.~\ref{fig:overview}.

\subsection{\issuefinder}
\label{issue finder}
\begin{figure}[t!]
  \centering
  \includegraphics[width=0.48\textwidth]{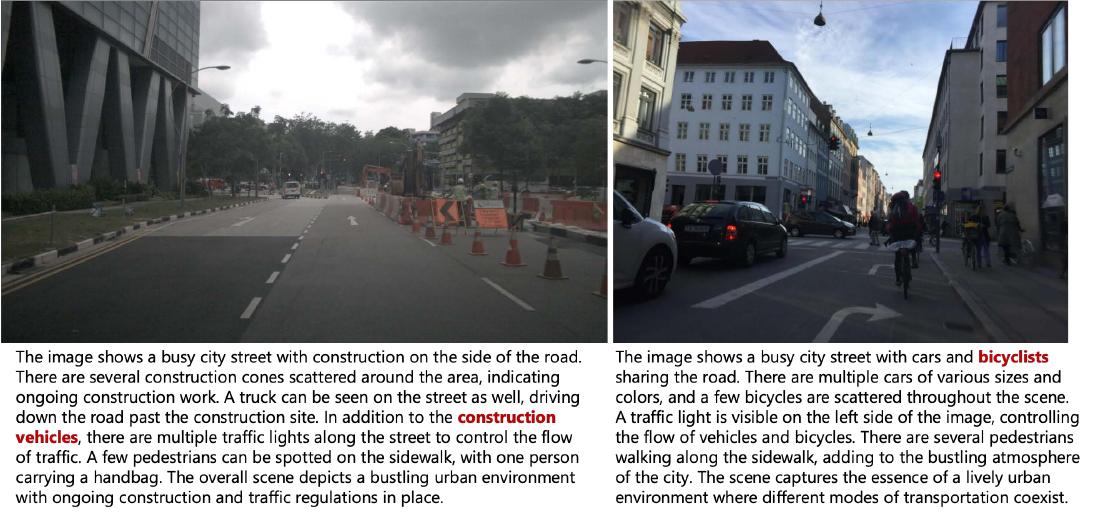}
  \caption{Examples of the \issuefinder. We use Otter~\cite{li2023otter} to generate detailed descriptions of an image, then identify the novel category that is missing in the label space (shown in red).}
  \label{fig:issue_finder}
\end{figure}

Given the large amount of unlabeled data collected by AVs in daily operation, identifying the missing category of existing label space is difficult as it requires humans to extensively compare the detection results and image context to spot the difference, which hinders the AV system's iterative development. To ease the difficulty, we consider the multi-modality dense captioning~(MMDC) models to automate the process. As the MMDC models like Otter~\cite{li2023otter} are trained with several million multi-modal in-context instruction tuning datasets, they can provide fine-grained and comprehensive descriptions of the scene context as shown in Fig.~\ref{fig:issue_finder}, and we conjecture that they may be more likely to return a synonym to the sought label of the novel category than an OVOD method to detect a bounding box for the novel category. Specifically, an unlabeled image will pass to both the detector deployed on-car and the MMDC model to get the list of predicted categories and the detailed captions of the image, respectively. By basic text processing, we can readily identify the novel category the model can not detect. In that case, our data engine will trigger the \datafeeder{} to query relevant images for incrementally training the detector to extend its label space correspondingly.   

\begin{figure}[t!]
  \centering
  \includegraphics[width=0.48\textwidth]{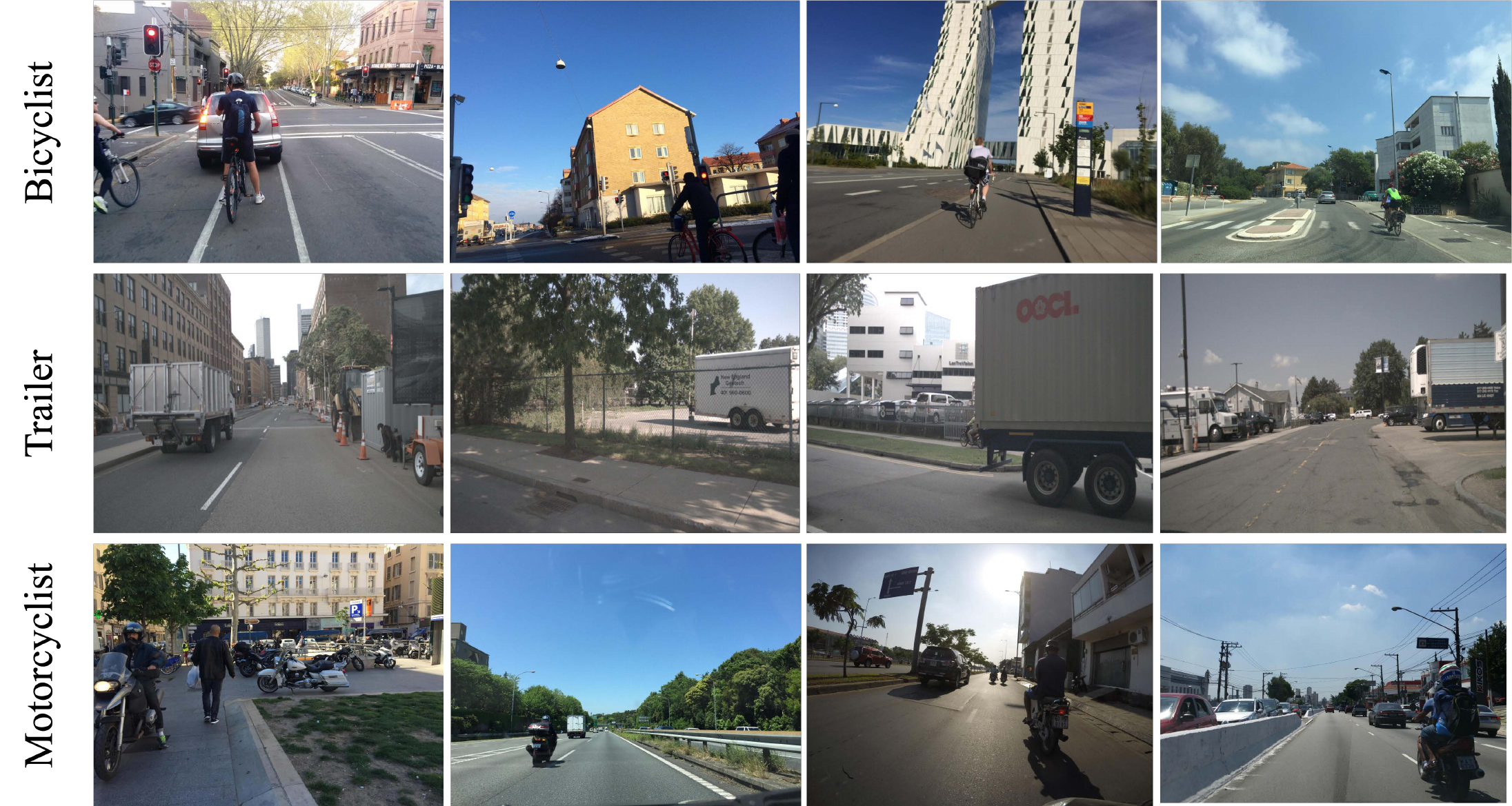}
  \caption{Visualization of the queried images from \datafeeder{} on three novel categories.} 
  \label{fig:data_feeder}
  \vspace{-5pt}
\end{figure}
\subsection{\datafeeder}

The purpose of \datafeeder{} is to first query meaningful images that could contain the novel category. The goal is to (1) reduce the search space for pseudo-labeling and accelerate pseudo-labeling in \modelupdater{}, and (2) remove trivial or unrelated images during training so we can reduce training time while also improving performance. 
This is especially important in real-world scenarios where a large amount of data can be collected every day.
As novel categories can be arbitrary and open-vocabulary, a naive solution is to search similar images like the input image of \issuefinder{} by exploiting the feature similarity, e.g., via similarity of the image feature by CLIP~\cite{radford2021learning}. However, we find that the image similarity cannot reliably identify sufficient numbers of relevant images due to the high variety of the AV datasets~(see Tab.~\ref{tab:ab_datafeeder}). Instead, our \datafeeder{} utilizes the VLMs to perform text-guided image retrieval on the image pool to query for relevant images related to the novel categories. We consider BLIP-2~\cite{li2023blip} given its strong open-vocabulary text-guided retrieval capability. Precisely, given an image and a specific text input, we measure the cosine similarity between their embeddings from BLIP-2 and only retrieve the top-$k$ images for further labeling in our \modelupdater. For the text prompt, we experiment with common prompt engineering practice~\cite{radford2021learning} and find that a template like \textit{``An image containing \{\}''} can readily provide good precision and recall for the novel categories in practice. Fig.~\ref{fig:data_feeder} shows some examples of retrieved images.

\begin{figure}[t!]
  \centering
  \includegraphics[width=0.48\textwidth]{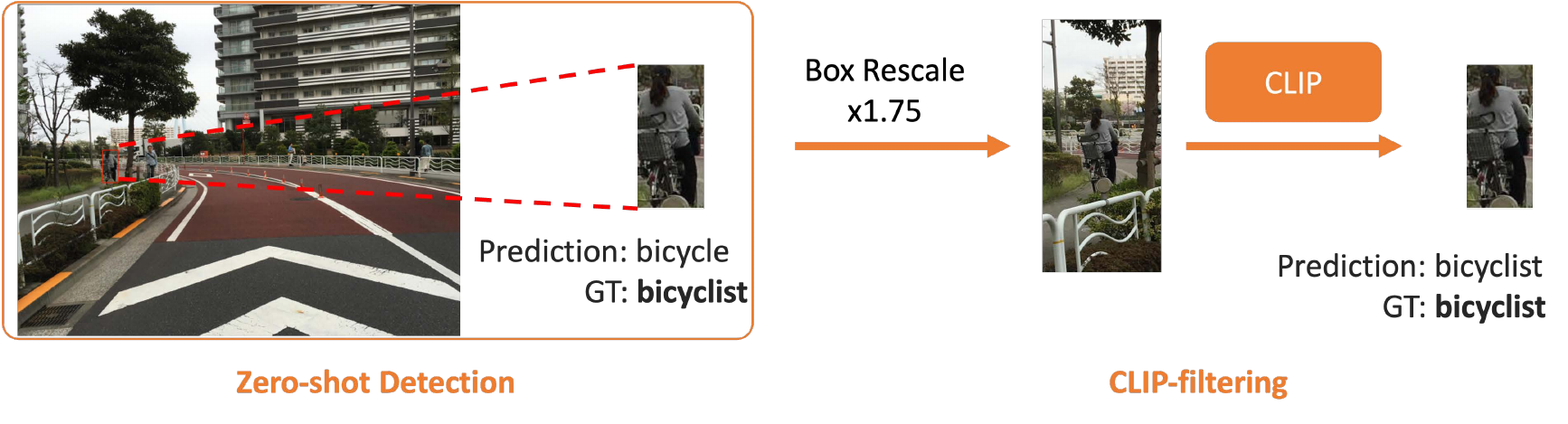}
  \caption{Our two-stage pseudo-labeling for \modelupdater{}: generate boxes by zero-shot detection and label by CLIP filtering.} 
  \label{fig:pseudo_label}
  \vspace{-5pt}
\end{figure}

\subsection{\modelupdater}
\label{model updater}
The goal of our \modelupdater{} is to make our detector learn to detect novel objects without human annotations. To this end, we perform pseudo-labeling on the images queried by the \datafeeder{} and then use them to train our detector. 

\subsubsection{Two-Stage Pseudo-Labeling}
\label{Two-Stage Pseudo-Labeling}
Motivated by the previous success in pseudo-labeling for object detection~\cite{zhao2022exploiting}, we designed our pseudo-labeling procedure with two parts: box and label generation. Such a two-stage framework can help us better dissect the issue of pseudo-label generation and improve the label generation quality. Box generation aims to identify as many object proposals in the image as possible, i.e., high recall for localizing novel categories, to guarantee a sufficient number of candidates for label generation. To this end, region proposal networks~(RPN) pretrained with closed-set label space~\cite{zhao2022exploiting} and the open vocabulary detectors~(OVD)~\cite{minderer2023scaling} can be considered, where the former can localize generic objects while the latter can perform text-guided localization. 
We observe that the SOTA OVD, i.e., OWL-v2~\cite{minderer2023scaling} that has been self-trained on web-scale datasets~\cite{chen2023pali}, exhibits a higher recall to localize novel categories compared to the RPN. We conjecture that proposals of RPN may be readily biased toward the pre-trained categories.

Thus, we choose OWL-v2 as our zero-shot detector to get the box proposal. Specifically, we append the novel category name provided by \issuefinder{} to our existing label space and create the text prompts, then we prompt the OWL-v2 to inference on an image. Note that we only retain the box proposals and remove the labels from the OWL-v2's predictions. This is because we empirically find that OWL-v2 can not achieve reliable precision on the novel categories presented in AV datasets, e.g., less than 10\% AP averaging over the novel categories in AV datasets~\cite{nuscenes2019,neuhold2017mapillary}, while it can get $>$40\% AP on novel categories of LVIS~\cite{gupta2019lvis} datasets. We conjecture that this performance degradation may come from the domain shift of the images collected in the AV scenario. For instance, the pretraining data of OWL-v2 mainly comes from the daily image captured by humans from a close distance. However, the street objects are always small in the image due to their long distance from the on-car camera, and the aspect ratio of the image presented in AV datasets is relatively large, making OWL-v2 hard to classify the correct label of the object proposals. 

Motivated by this insight, we consider conducting another round of label filtering with CLIP~\cite{radford2021learning} to purify the predictions of the OWL-v2 and generate the pseudo labels. Specifically, we pass the box prediction by OWL-v2 to the original CLIP model~\cite{radford2021learning} for zero-shot classification~(ZSC), as shown in Fig.~\ref{fig:pseudo_label}. To mitigate the potential issue of the aspect ratio mentioned above, we increase the box size to crop the image and then send the cropped image patch to CLIP for ZSC. This can involve more scene contextual information to help the CLIP better differentiate between the novel and known categories. Regarding the label space for CLIP to do zero-shot classification, we first create a base label space, which is a combination of the label space from datasets we have pre-trained and COCO~\cite{lin2014microsoft}, to ensure that we can mostly cover daily objects that would probably be present in the street. 
The base label space will automatically extend when the \issuefinder{} identifies novel categories not in the base label space.

\subsubsection{Continual Training with Pseudo-labels}
\label{Continual Training with Pseudo-labels}

Directly training our existing detector on the pseudo-labels of novel categories presents a challenge, as these labels may lead the detector to overfit and catastrophically forget the known categories. The issue arises because the unlabeled data can contain both novel and known categories that the detector has previously learned. Without labels for those known categories and only having labels for novel categories, the model may incorrectly suppress predictions for known categories, focusing solely on predicting novel categories. As training progresses, the known categories gradually fade from memory.
To address this issue, we draw inspiration from existing self-training strategies and include the pseudo-labels of the known categories that have been trained on. Consequently, our existing detector is updated with the pseudo-labels of both novel and known categories. To obtain pseudo-labels for the known categories, we first use our detector to infer data before applying OWL-v2 to the data. Empirically, we find that including pseudo-labels for known categories helps the model distinguish between known and novel categories, boosting the performance of novel categories and mitigating the catastrophic forgetting issues associated with known categories.
Additionally, acknowledging that pseudo-labels for both known and novel categories may not be perfect, we filter the pseudo-labels. 
For known categories, we only use pseudo-labels with high predicted confidence from our detector. For novel categories, we have already incorporated CLIP to filter pseudo-labels, as mentioned in Section~\ref{Two-Stage Pseudo-Labeling}.
\begin{figure}[t!]
  \centering
  \includegraphics[width=0.5\textwidth]{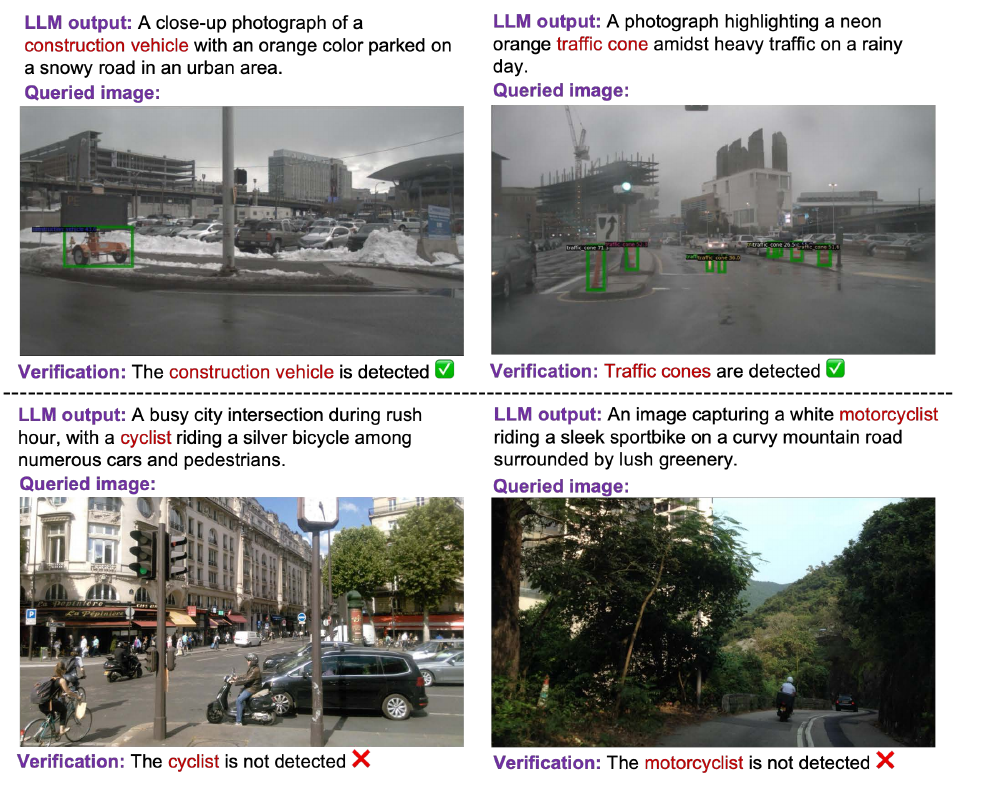}
  \caption{Visualization on the \verification. \textbf{\color{Purple}{LLM output}}: We use LLM to generate descriptions of the novel category with variations of the scenarios. \textbf{\color{Purple}{Queried image}}: For each description, we use VLM to query images from our training data. \textbf{\color{Purple}{Verification}}: we let humans review whether the novel category has been detected.}
  \label{fig:verification}
  \vspace{-10pt}
\end{figure}
\begin{table*}[ht!]
  \centering
  \footnotesize	
    \begin{tabular}{c|c|cc|ccc}
    \toprule
    \multirow{2}[1]{*}{Method} & \multirow{2}[1]{*}{Algorithm} & \multicolumn{2}{c|}{Cost (\$)} & \multicolumn{3}{c}{Accuracy (\%)} \\
    & & Training & Labeling & \multicolumn{1}{c}{Novel} & \multicolumn{1}{c}{Known} & \multicolumn{1}{c}{Forgetting} \\
    \midrule
    Fully-Supervised &  &  0.3     &  1005.2     &   24.1    &  29.9     &  - \\
    \midrule
    \multirow{2}[1]{*}{Open Vocabulary Object Detection} & OwL-ViT~\cite{minderer2022simple} &   0.9    &   0    &  2.0     &  5.5     &  - \\ 
          & OwL-v2~\cite{minderer2023scaling}  &  0.9     &   0    &   9.7    &   17.9   & - \\ 
    \midrule
    Semi-Supervised Learning &  Unbiased Teacher-v1~\cite{liu2021unbiased} &   1.1    &   1.0 &   6.3   &  1.2   & -28.7  \\
    \midrule

    \multirow{2}[1]{*}{AIDE (Ours)} & w/o \datafeeder{}  &   5.7    &   0    &   10.1    &  26.8   &  -3.1  \\

     & w/ \datafeeder{}  &  0.6    &   0   &   12.0    &  26.6    & -3.3  \\
    \bottomrule
    \end{tabular}%
  \caption{Cost and accuracy for fully-supervised, open-vocabulary object detection, semi-supervised learning, and our data engine (AIDE) to detect one novel category from Mapillary and nuImages. We initialize Semi-SL and ours with the same detector.}
  \label{tab:main_table_avg_only}%
\end{table*}%

\begin{table*}[ht!]
  \centering
  \footnotesize	
  \begin{tabular}{cc|c|cccc|c|cc}
    \toprule
    \multicolumn{2}{c|}{Method $\longrightarrow$} & OVOD & \multicolumn{4}{c|}{Supervised Training} & Semi-SL & \multicolumn{2}{c}{AIDE (Ours)} \\
    \multicolumn{2}{c|}{Algorithm $\longrightarrow$} & OWL-v2~\cite{minderer2023scaling} & \multicolumn{4}{c|}{} &  UTeacher-v1~\cite{liu2021unbiased} & w/o \datafeeder{} & w/ \datafeeder{} \\
    \midrule
    \multicolumn{2}{c|}{\#Labels per Category $\longrightarrow$} & 0 & 10 & 20 & 50 & All & 10 & 0 & 0 \\
    \midrule
    Mapillary & motorcyclist & 4.0 & 5.9 & 12.4 & 13.7 & 19.6 & 8.3 & 4.0 & 8.4 \\
    Mapillary & bicyclist & 0.9 & 8.9 & 10.8 & 12.4 & 22.4 & 3.5 & 7.7 & 11.9 \\
    nuImages & construction vehicle & 4.7 & 3.4 & 8.4 & 7.3 & 22.6 & 4.3 & 5.4 & 5.7 \\
    nuImages & trailer & 3.6 & 0.3 & 1.3 & 1.9 & 13.6 & 0.4 & 2.2 & 3.7 \\
    nuImages & traffic cone & 35.3 & 12.9 & 21.4 & 28.5 & 42.2 & 16.4 & 31.0 & 30.7 \\
    \midrule
    \multicolumn{2}{c|}{Average} & 9.7 & 6.3 & 10.9 & 12.8 & 24.1 & 6.6 & 10.1 & 12.0 \\
    \bottomrule
  \end{tabular}
  \caption{Per-category accuracy (AP \%) on novel categories with different methods.}
  \label{tab:main_table_per_class}
  \vspace{-10pt}
\end{table*}
\subsection{\verification}
The \verification{} step aims to evaluate whether the updated detector can detect the novel categories under different scenarios, to ensure the model can handle unexpected or unseen scenarios.
To this end, we prompt the ChatGPT~\cite{ChatGPT} with the name of novel categories to generate diverse scene descriptions. 
These descriptions contain variations of the scenarios, such as different appearances of the objects, surrounding objects, time of the day, weather conditions, etc.
For each scene description, we again use BLIP-2 to query relevant images, which are used to test the model's robustness. To ensure the correctness, we ask humans to review if the predictions for the novel categories are correct. 
If the predictions are correct, the detector has passed the unit test. 
Otherwise, we ask humans to provide the ground-truth label, which can be used to further improve the model.
Compared to existing solutions that have humans manually examine the model prediction one by one, our \verification{} exploits the LLM to facilitate the search for potential failure cases by diverse scene generation, where the search cost can be largely saved, and the cost of verifying a correct detection or even fixing an incorrect one is lower.

\section{Experiments}

\subsection{Experimental Setting} 

\textbf{Datasets and Novel Categories Selection} In reality, the AV system can hardly train with a single source of data, e.g., AVs may operate in various locations in the world to collect data. 
To simulate such a nature faithfully, we leverage the existing AV datasets to jointly train our closed-set detector, including Mapillary~\cite{neuhold2017mapillary}, Cityscapes~\cite{Cordts2016Cityscapes}, nuImages~\cite{nuscenes2019}, BDD100k~\cite{bdd100k}, Waymo~\cite{Sun_2020_CVPR}, and KITTI~\cite{Geiger2012CVPR}. 
We use this pretrained detector as the initialization for the supervised training, Semi-SL, and our AIDE for a fair comparison.
There are 46 categories in total after combining the label spaces. 
To simulate the novel categories and ensure that the selected categories are meaningful and crucial for AV in the street, we choose 5 categories as novel categories: ``motorcyclist'' and ``bicyclist'' from Mapillary, ``construction vehicle'', ``trailer'', and ``traffic cone'' from nuImages.
The rest 41 categories are set as known.
We remove all the annotations for these categories in our joint datasets and also remove the related categories with similar semantic meanings, e.g., ``bicyclist'' vs ``cyclist''. We attach more details of the dataset statistics in the supplementary material.

\noindent{\textbf{Methods for Comparison}} To our knowledge, there is little work about the systematic design for automatic data engines tailored to the novel object detection for AV systems. Thus, it is hard to identify a comparable counterpart for our AIDE. To this end, we dissect our evaluation into two parts: (1) compare to alternative detection methods and learning paradigms on the performance of novel object detection; (2) ablation study and analysis of each step of the automatic data engine. For (1), as our AIDE can enable the detector to detect novel categories without any labels, we first compare our method with the zero-shot OVOD methods on novel categories' performance. Moreover, to show the efficiency and effectiveness of our AIDE in reducing label cost, we further compare with semi-supervised learning~(Semi-SL) and fully supervised learning that trains the detector with different ratios of ground-truth labels. Specifically, we compare our data engine to state-of-the-art~(SOTA) OVOD methods like OWL-v2~\cite{minderer2023scaling}, OWL-ViT~\cite{minderer2022simple}, and Semi-SL methods like Unbiased Teacher~\cite{liu2021unbiased, liu2022unbiased}.

\noindent{\textbf{Experimental Protocols}} We treat each of the five selected classes as novel classes and conduct experiments separately to simulate the scenario that one novel class has been identified at a time by our \issuefinder. For Semi-SL methods, we provide different numbers of ground-truth images for training. 
Each image could contain one or multiple objects of the novel category. 
We evaluate all comparison methods on the dataset of the novel category for a fair comparison.

\noindent{\textbf{Evaluation}} As our AIDE automates the whole data curation, model training, and verification process for the AV system, we are interested in how our engine can strike a balance between the cost of searching and labeling images and the performance on novel object detection. 
We measure the human labeling costs~\cite{adhikari2018faster} and also the GPU inference costs~\cite{Lambda}, i.e., the usage of VLMs/LLMs in our AIDE and training the model with pseudo labeled for our AIDE or with ground-truth labels for comparison methods, denoted as `Labeling + Training Cost' in Fig.~\ref{fig:teaser}. 
The labeling cost for a bounding box is \$0.06~\cite{adhikari2018faster}, and the GPU cost is \$1.1 per hour~\cite{Lambda}. The cost of ChatGPT is negligible ($<$ \$0.01).

\noindent{\textbf{Experimental Details}} Given the real-time requirement for inference, we choose the Fast-RCNN~\cite{girshick2015fast} as our detector instead of OVOD methods like OWL-ViT~\cite{minderer2022simple} as the FPS for OWL-ViT is only 3. We run our AIDE to iteratively scale up its capability of detecting novel objects. For multi-dataset training, we follow the same recipe from~\cite{zhao2020object}. For each novel category, we train for 3000 iterations with the learning rate of 5e-4, and we use the same hyperparameter for all the comparison methods if they require training. We attach our full experimental details in the supplementary material.
 
\subsection{Overall Performance}

In this section, we provide the overall performance of novel object detection after running our AIDE for a complete cycle. 
Our results are shown in Fig.~\ref{fig:teaser} and Tab.~\ref{tab:main_table_avg_only}.
Compared to the SOTA OVOD method, OwL-v2~\cite{minderer2023scaling}, our method outperforms by 2.3\%AP on novel categories and 8.7\%AP on known categories, showing that our AIDE can benefit from mining the open-vocabulary knowledge from OVOD method.
This is due to our simple yet effective continual training strategy described in Section~\ref{Continual Training with Pseudo-labels}. 
Moreover, our AIDE suffers much less from catastrophic forgetting compared to Semi-SL methods, since current Semi-SL methods for object detection do not contain continual learning settings. 
Existing works on continual semi-supervised learning~\cite{wang2021ordisco,kang2023soft} only consider image classification and are not applicable to object detection. Combining our AIDE with and without the \datafeeder{} makes it apparent that our \datafeeder{} can sufficiently reduce the inference time cost as the \datafeeder{} can pre-filter irrelevant images, and the \modelupdater{} only needs to assign pseudo-labels on a small number of relevant images. 
Tab.~\ref{tab:main_table_avg_only} shows that pre-filtering leads to better AP on novel categories. 

\subsection{Analysis on AIDE}
In the following subsections, we will dissect each part of our AIDE to validate our design choice.
\subsubsection{\issuefinder}

As mentioned in Section~\ref{issue finder}, the main goal of our \issuefinder{} is to automatically identify categories that do not exist in our label space. 
To this end, we evaluate the success rate of automatically identifying the novel categories.
We find that dense captioning models can automatically predict if the image contains the novel categories more precisely, compared to using OVOD methods to identify and localize novel objects when they are given the names of the novel categories, as shown in Tab.~\ref{tab:ab_issuefinder}. 
Note that the goal here is to only identify the missing categories, hence we choose to use dense captions here and leverage OVOD to help localize the novel object in the later steps.

\begin{table}[t!]
    \centering
    \footnotesize
    \begin{tabular}{cccc}
        \toprule
        \multirow{2}[1]{*}{Dataset} & \multirow{2}[1]{*}{Category Name} & Dense Captioning & OVOD \\
        & & Precision (\%) & AP50 (\%)\\
        \midrule
        Mapillary & motorcyclist & 83.3 & 9.5 \\
        Mapillary & bicyclist & 89.5 & 1.6 \\
        nuImages & const. vehicle & 65.6 & 12.9 \\
        nuImages &trailer & 24.7 & 7.1 \\
        nuImages &traffic cone & 87.9 & 60.3 \\
        \midrule
        \multicolumn{2}{c}{Average} & 70.2 & 18.3\\
        \bottomrule
    \end{tabular}
    \caption{Comparing with using OVOD to identify and localize novel categories, Dense Captioning better predicts missing categories more reliably in our \issuefinder{}.}
    \label{tab:ab_issuefinder}
    \vspace{-5pt}
\end{table}

\subsubsection{\datafeeder{}}
\begin{table}[t!]
\footnotesize
  \centering
    \begin{tabular}{ccccc}
    \toprule
    \multirow{2}[1]{*}{Dataset} & \multirow{2}[1]{*}{Category} & \multirow{2}[1]{*}{Image similarity} & \multicolumn{2}{c}{VLM Retrieval} \\
     & & & CLIP & BLIP-2 \\
    \midrule
    Mapillary & motorcyclist & 22.6    & 19.0    & 50.4 \\
    Mapillary & bicyclist & 17.9    & 28.8    & 50.5 \\
    nuImages & const. vehicle & 14.2    & 51.2    & 55.6 \\
    nuImages & trailer & 10.5    & 23.3    & 16.5 \\
    nuImages & traffic cone & 29.5    & 47.3    & 99.3 \\
    \midrule
    \multicolumn{2}{c}{Average} & 18.9 & 33.9 & 54.5 \\
    \bottomrule
    \end{tabular}%
    \caption{Ablation studies of the \datafeeder. We report accuracy~(\%) of the top-1$k$ images queried by image similarity search and text-based retrieval with VLM, i.e., CLIP and BLIP-2.}
  \label{tab:ab_datafeeder}%
  \vspace{-5pt}
\end{table}

The goal of the \datafeeder{} is to curate relevant data from a large pool of images with high precision.
We compare several choices, including image similarity search by CLIP feature, and text-guided image retrieval by VLMs, i.e., BLIP-2 and the CLIP. 
We report the accuracy of top-$k$ queried images over different categories in Tab.~\ref{tab:ab_datafeeder}, showing that image similarity search is inferior to VLMs. 
This is because the novel categories can have large intra-class variations, and thus only one image may not be representative of finding sufficient amounts of relevant images. 
Compared with CLIP, our choice of BLIP-2 performs better on average. 

\subsubsection{\modelupdater}
\label{ablation study for model updater}
We ablate the design choices for our box and pseudo-label generation. 
For box generation, we compare our choice of using box proposals from OWL-v2 with using proposals from VL-PLM~\cite{zhao2022exploiting}, which generates box proposals by the region proposal network~(RPN) of MaskRCNN~\cite{he2017mask} pretrained on COCO. 
We also compare with using proposals from Segment Anything model~(SAM)~\cite{kirillov2023segment}, specifically we use the FastSAM~\cite{zhao2023fast} since it is faster in inference while having the same performance as SAM.
As shown in the ablation studies in Tab.~\ref{tab:ab_modelupdater}, our choice of using OWL-v2 is the best among using VL-PLM and SAM. We observe that SAM may generate many small objects with no semantic meaning, suppressing the effective amount of pseudo-labels. This is expected as the pre-training of SAM does not use semantic labels. 
For label generation, we compare with using OWL-v2 prediction directly without filtering by CLIP, i.e., ``w/o CLIP'', showing that filtering labels with CLIP is necessary. 
Last, compared with training our detector without pseudo-labels of known category, denoted as ``ex.~known'', we outperform by 3.9\% AP on novel categories. 
Moreover, the AP of known categories without using pseudo-label is only 1.58\%, while Ours is 26.6\% as shown in Tab.~\ref{tab:main_table_avg_only}. 
This verifies the effect of using pseudo-labels of known categories as discussed in Sec.~\ref{Continual Training with Pseudo-labels}.

\begin{table}[t!]
    \centering
    \footnotesize
    \begin{tabular}{cccccc}
        \toprule
        Category & SAM & VL-PLM & w/o CLIP & \scriptsize{ex.~known}& Ours \\
        \midrule
        motorcyclist & 0.5 & 10.1 & 3.3 & 2.8 & 8.4 \\
        bicyclist & 2.8 & 6.5 & 3.2 & 2.1 & 11.9 \\
        const. vehicle & 1.4 & 4.3 & 4.0 & 3.5 & 5.7 \\
        trailer & 0.4 & 0.4 & 2.0 & 1.1 & 3.7 \\
        traffic cone & 14.5 & 10.4 & 30.0 & 30.9 & 30.7 \\
        \midrule
        Average AP~(\%) & 3.9 & 6.3 & 8.5 & 8.1 & 12.0 \\
        \bottomrule
    \end{tabular}
    \caption{Ablation of \modelupdater{} on box generation with SAM and VL-PLM, label generation without CLIP filtering, and continual training excluded pseudo labels of known categories.}
    \label{tab:ab_modelupdater}
    \vspace{-5pt}
\end{table}

\begin{table}[t!]
\footnotesize
  \centering
    \begin{tabular}{ccc}
    \toprule
    {Dataset} & {Category} & Diversity (\%) \\
    \midrule
    Mapillary & motorcyclist  & 57.6  \\
    Mapillary & bicyclist  & 62.2  \\
    nuImages & const. vehicle  & 77.0   \\
    nuImages & trailer  & 82.0  \\
    nuImages & traffic cone & 70.4 \\
    \midrule
    \multicolumn{2}{c}{Average} &  69.8 \\
    \bottomrule
    \end{tabular}%
    \caption{Our \verification{} step can indeed find diverse scenarios. The diversity is measured by the number of distinct images among 100 queried images using descriptions generated by ChatGPT. }
    \label{tab:ab_verification}%
    \vspace{-10pt}
\end{table}%

\subsubsection{\verification}
\label{verification experiment}
The goal of the \verification{} is to evaluate the detector's robustness and to verify the performance under diverse scenarios. 
Humans only need to examine if the predictions are correct in each scenario which reduces the monitoring cost since the scenarios are diverse and it takes less time to check the predictions than to annotate. To test if the generated scenarios are diverse, we measure the number of unique images among 100 images queried by generated descriptions and repeat the process ten times. 
As shown in Tab.~\ref{tab:ab_verification}, our \verification{} can indeed find diverse scenarios, as 69.8\% images are distinct on average, even on such small training datasets.

\begin{figure}[t!]
  \centering
  \includegraphics[width=0.5\textwidth]{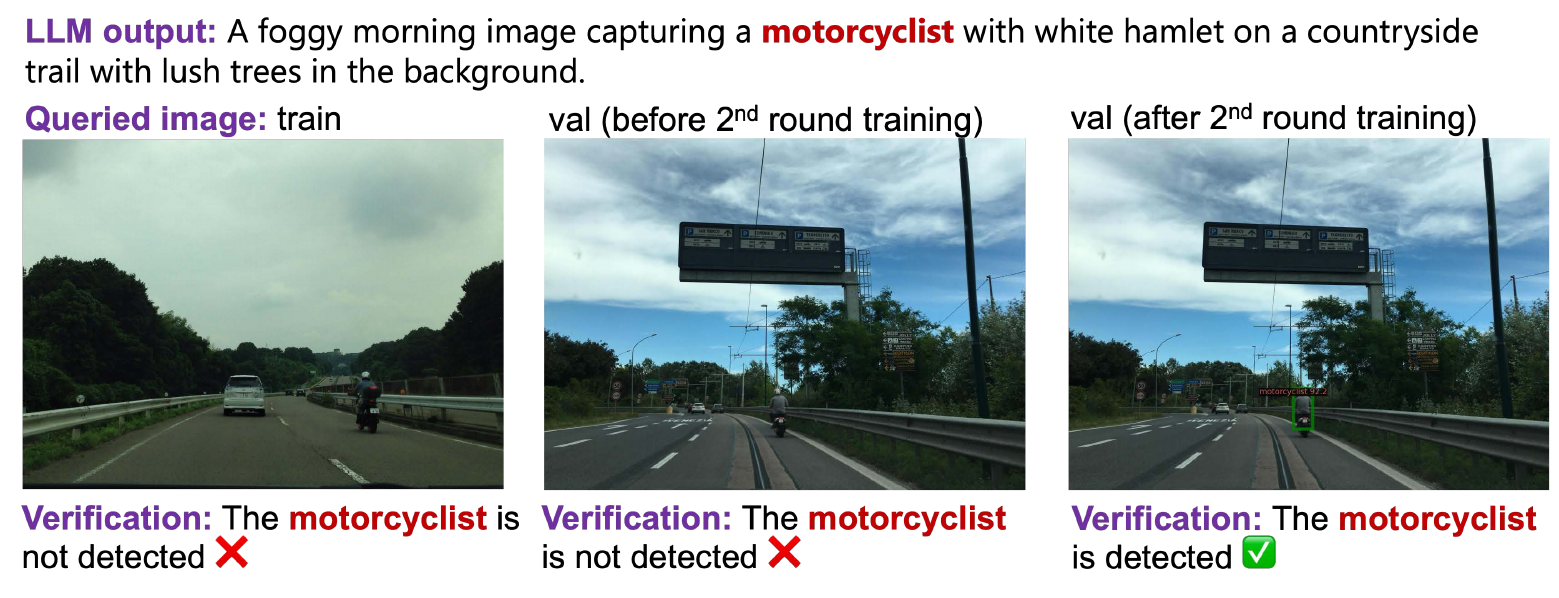}
  \caption{Visualization on the \verification. 
  \textbf{Left:} In the queried image from the training set for verification, the model is not predicting the motorcyclist.
  \textbf{Middle:} Similarly on the queried image from the validation set, the model is not predicting the motorcyclist.
  \textbf{Right:} After updating the model again, our model can successfully predict the motorcyclist. 
  }
  \label{fig:verification_round2}
  \vspace{-10pt}
\end{figure}

If the prediction is incorrect, we can ask annotators to label the images, which are used to further improve the detector. To this end, we randomly select 10 LLM-generated descriptions, for which top-1 retrieved image~(based on BLIP-2 cosine similarity) was predicted incorrectly, and labeled these 10 images to update our detector by \modelupdater{}.
As shown in Fig.~\ref{fig:verification_round2}, after updating the model with a few human supervisions, our model can successfully predict the object, e.g., the motorcyclist in the figure, which was miss-detected before. 
For the overall performance, we achieve 14.2\% AP on novel categories, which improves our zero-shot performance by 2.2\% AP, while the total cost only increases to \$1.59. 
This is still less than \$2.1 of semi-supervised learning, and our AP for known categories remains 26.6\% after \verification{}. 

\section{Conclusion}
We proposed an Automatic Data Engine (AIDE) that can automatically identify the issues, efficiently curate data, improve the model using auto-labeling, and verify the model through generated diverse scenarios. By leveraging VLMs and LLMs, our pipeline reduces labeling and training costs while achieving better accuracies on novel object detection. The process operates iteratively which allows continuous improvement of the model, which is critical for autonomous driving systems to handle expected events. We also establish a benchmark for open-world detection on AV datasets, demonstrating our method's better performance at a reduced cost. 
One of the limitations of AIDE is that VLM and LLM can hallucinate in issue finder and verification. Despite the effectiveness of AIDE, for a safety-critical system, some human oversight is always recommended.

\textbf{Acknowledgements} This work was supported in part by National Science Foundation grant IIS-2007613.

{
    \small
    \bibliographystyle{unsrt}
    \bibliography{main}
}

\clearpage
\setcounter{page}{1}
\maketitlesupplementary
\appendix

\section{\verification{} can Boost AIDE's Performance}

In \verification{}, humans are asked to verify the predictions on the diverse scenarios generated by LLMs (ChatGPT~\cite{ChatGPT}).
If the prediction is incorrect, annotators can give correct bounding boxes, which can be used by AIDE to self-improve the model.
In this section, we examine whether these annotations can boost the performance of AIDE. 
To this end, we train the model after we have collected annotations for 10, 20, and 30 images.
However, since we only have a few human annotations collected, directly combining them with a large number of pseudo-labels from the \modelupdater{} will cause issues if we have a uniform sampling rate on the data loader during training.

On the other hand, semi-supervised learning methods like Unbiased Teacher-v1~\cite{liu2021unbiased} have demonstrated notable performance on novel categories with minimal annotations, owing to their strong augmentation strategy.

\begin{figure}[t]
  \centering
  \includegraphics[width=0.45\textwidth]{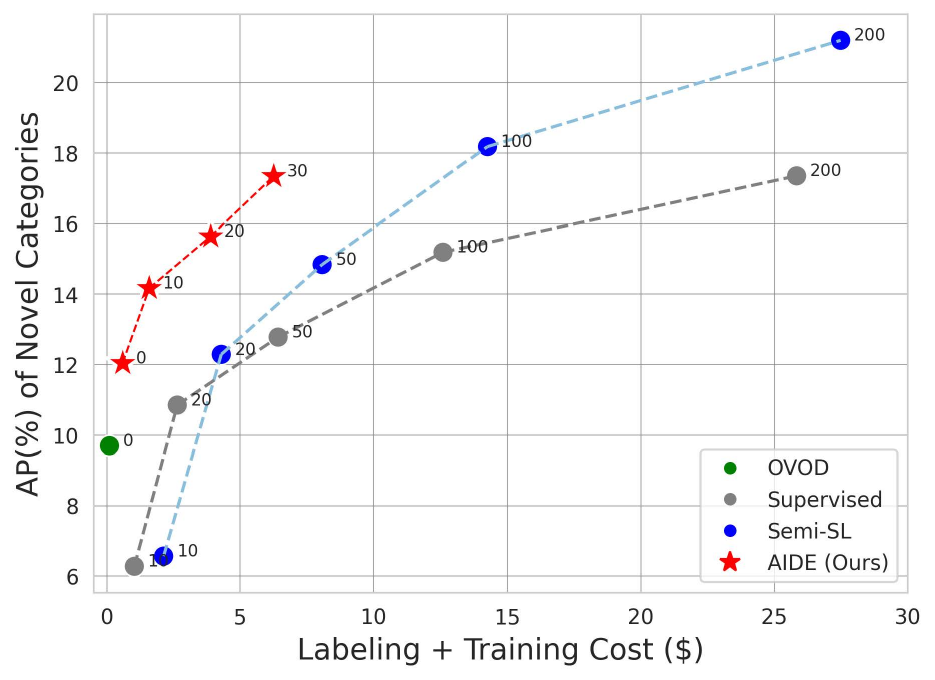}
  \caption{We demonstrate that the annotations in the \verification{} step can boost the performance of AIDE. 
  The numbers next to the data points denote the number of labeled images used by each method. 
  Note that AIDE only introduces labeled images in \verification{} if an annotator wants to provide the labels when the detector gives incorrect predictions on the test scenarios. 
  }
  \label{fig:more verification}
\end{figure}

Motivated by this insight, we first use the few labeled images to train an auxiliary model by the strong augmentation strategy as \cite{liu2021unbiased} but with 1000 iterations to reduce training costs. This auxiliary model is then used to generate pseudo-labels for the novel categories based on the images initially queried by our \datafeeder{}, and these are combined with the earlier pseudo-labels generated by our \modelupdater{} for both novel and known categories to fine-tune our detector again in our \modelupdater{}. By doing so, we can obtain more pseudo-labels for novel categories with high quality and alleviate the sampling issue in the data loader. 
As shown in Fig.~\ref{fig:more verification}, our AIDE can be largely improved.

\section{More Comparisons between AIDE and OVOD (OWL-v2)}
\label{sec: update AIDE}
In this section, we demonstrate that AIDE is a general automatic data engine that can enhance different object detectors for novel object detection. 
Specifically, we replace the closed-set detector (Faster RCNN~\cite{ren2015faster}) with the state-of-the-art (SOTA) open-vocabulary object detection (OVOD) method, OWL-v2.

As shown in Tab.~\ref{tab:ab_owlv2}, by applying our AIDE on OWL-v2, we can achieve 13.2\% AP on average without human annotations, marking a 3.5\% improvement over the original OWL-v2 model.
However, our default detector is Faster RCNN since it has a faster inference speed, which is favorable for autonomous driving.

In addition, the original OWL-v2 paper~\cite{minderer2023scaling} proposes a self-training strategy to enhance the OWL-v2 on novel object detection, i.e., directly using the predictions of OWL-v2 with a certain confidence threshold to self-train the OWL-v2. We compare this self-training schedule with our AIDE.

As shown in Tab.~\ref{tab:ab_owlv2}, the self-training can improve the OWL-v2, but it is still inferior to AIDE 3.1\%. 
This improvement is attributable to our \datafeeder{} and the CLIP filtering in our \modelupdater{}, which help to minimize irrelevant images for pseudo-labeling and filter out inaccurate OWL-v2 predictions, thereby enhancing the quality of pseudo-labels and the subsequent performance after fine-tuning OWL-v2 with these labels. 
We will dissect the impact of our \datafeeder{} and \modelupdater{} on improving the quality of pseudo-label in Sec.~\ref{dissect the impact} and Tab.~\ref{tab:disect_datafeeder_model_updater}. 

\begin{table}[t]
    \centering
    \footnotesize
    \begin{tabular}{c|cc|cc}
        \toprule
        \multirow{2}{*}{Categoty}& \multicolumn{2}{c|}{OVOD} & \multicolumn{2}{c}{AIDE (Ours)} \\
    & OWL-v2 & OWL-v2 ST & Faster RCNN & OWL-v2 \\
        \midrule
        motorcyclist & 4.0 & 5.3 & 8.4 & 11.4\\
        bicyclist & 0.9 & 0.8 & 11.9 & 9.8\\
        const. vehicle & 4.7 & 5.4 & 5.7 & 6.0\\
        trailer & 3.6 & 3.5 & 3.6 & 3.6 \\
        traffic cone & 35.3 & 35.5 & 30.7 & 35.3\\
        \midrule
        Average AP(\%) & 9.7 & 10.1 & 12.0 & 13.2\\
        \bottomrule
    \end{tabular}
    \caption{Comparison between OWL-v2, OWL-v2 with self-training, and AIDE on improving an existing detector on novel object detection with any human annotations. ST: Self-training using the same strategy in~\cite{minderer2023scaling}.
    }
    \label{tab:ab_owlv2}
\end{table}

\begin{table*}[t!]
    \centering
    \small
    \begin{tabular}{cccccc}
        \toprule
        Dataset & Category & Mapillary / nuImages & +Waymo~(39k) & +Waymo~(78k) & +Waymo~(78k) +BDD100k~(69k)  \\
        \midrule
        Mapillary & motorcyclist & 8.4  &  9.4 & 11.1& 13.4\\
        Mapillary & bicyclist & 11.9  &  13.0 & 15.0& 18.4\\
        nuImages & const. vehicle & 5.7  & 7.3 & 14.6& 19.7\\
        nuImages &trailer & 3.7 & 3.6  & 5.1& 11.2\\
        nuImages &traffic cone & 30.7 & 31.6 & 35.1&  36.1\\
        \midrule
        \multicolumn{2}{c}{Average AP(\%)} & 12.0  & 13.9 & 16.2& 19.8\\
        \bottomrule
    \end{tabular}
    \caption{Extending the image pool with the Waymo and BDD100k dataset in \datafeeder{} can boost the performance of AIDE.}
    \label{tab:more_waymo}
\end{table*}

\section{Extending the Image Pool further boosts AIDE's Performance}
\label{sec: extend with Waymo}
Our \datafeeder{} queries images from either Mapillary~\cite{neuhold2017mapillary} or nuImages~\cite{nuscenes2019} by default. 
To verify the scalability of AIDE, we add the Waymo dataset in the database for \datafeeder{}, i.e., the image pool for querying becomes \{nuImages,~Waymo\} or \{Mapillary,~Waymo\} for each novel category. 
Note that the Waymo dataset only contains three coarse labels, i.e., ``vehicle'', ``pedestrian'', and ``cyclist'', as shown in Tab.~\ref{tab:datasets_labelspace}. 
Therefore it is uncertain whether novel categories such as ``motorcyclist'', ``construction vehicle'', 
``trailer'', and  ``traffic cone'' are present in the Waymo dataset.
For ``bicyclist'', although the Waymo dataset includes a similar label ``cyclist'', we have excluded all annotations of this category as described in Sec.~4.1 of our main paper. 
Moreover, given that the Waymo dataset consists largely of videos, resulting in numerous similar images, we implemented a sampling strategy. Each video was subsampled with a frame rate of 20, reducing the total number of images from 790,405 to 39,750~(denoted as 39k). 
We used the same hyperparameters for BLIP-2 and CLIP in our \datafeeder{} and \modelupdater{} as were used for the Mapillary and nuImages datasets, respectively, for image querying and pseudo-labeling.

As indicated in Table~\ref{tab:more_waymo}, incorporating the Waymo dataset into our \datafeeder{} for image querying resulted in a 1.9\% AP improvement in detecting novel categories, compared to using only the Mapillary or nuImages datasets. Moreover, by adding more unlabeled images from Waymo and the full BDD100k dataset, we can boost the performance to 19.8\% AP, approaching the fully-superivsed result of 24.1\% AP. Note that the cost of AIDE is only \$2.4 with 19.8\% AP. This significant improvement demonstrates that our AIDE can effectively scale up with an expanded image search space. 

\section{More Analysis}

\subsection{Ablation Study of the Scaling Ratio for CLIP filtering}
 
As discussed and illustrated in Sec.~3.3.1 and Fig.~5 of our main paper, we increase the size of the pseudo-box used to crop the image before submitting the cropped image patch for zero-shot classification (ZSC). 
We present an ablation study of the scaling ratio, ranging from 1.0 to 2.0, where a scaling ratio of 1.0 signifies using the pseudo-box dimensions as they are to crop the image patch. 
As Table~\ref{tab:ab_scaling_ratio} demonstrates, the performance of novel categories improves as the scaling ratio increases, reaching a plateau when the scaling ratio is 1.75. This trend is expected since a substantially rescaled box might include excessive background context, potentially distracting the ZSC process of CLIP. Therefore, we use a scaling ratio of 1.75 for all our experiments.

\begin{table}[t!]
  \centering
    \footnotesize
    \begin{tabular}{ccccccc}
    \toprule
    \multirow{2}{*}{{Dataset}} & \multirow{2}{*}{{Category Name}}  & \multicolumn{5}{c}{Scaling Ratio} \\
    & & 1     & 1.25  & 1.5   & 1.75  & 2 \\
    \midrule
    Mapillary & motorcyclist & 3.6   & 6.1   & 7.6   & 8.4   & 8.9 \\
    Mapillary & bicyclist & 9.3   & 10.7  & 12.0  & 11.9  & 12.2 \\
    nuImages & cons. vehicle & 5.8   & 5.0   & 4.8   & 5.7   & 5.4 \\
    nuImages & trailer & 2.1   & 2.1   & 3.2   & 3.6   & 3.6 \\
    nuImages & traffic cone & 28.6  & 30.2  & 28.6  & 30.7  & 29.2 \\
    \midrule
    \multicolumn{2}{c}{Average AP(\%)} & 9.9   & 10.8  & 11.2  & \textbf{12.0} & 11.8 \\
    \bottomrule
    \end{tabular}%
    \caption{Ablation study of the scaling ratio of the pseudo-box to crop the image patch for CLIP filtering.}
  \label{tab:ab_scaling_ratio}%
\end{table}%

\subsection{Analyzing the \datafeeder{} and \modelupdater{} on Improving the Quality of Pseudo-labeling}
\label{dissect the impact}

We analyze the impact of our \datafeeder{} and \modelupdater{} on improving the quality of pseudo-labels. As outlined in Section 3.2 of our main paper, our \datafeeder{} is designed to query images relevant to novel categories from the image pool. 
This process helps eliminate trivial or unrelated images during training, thereby reducing training time and enhancing performance. 
Moreover, our two-stage pseudo-labeling in our \modelupdater{} will filter out raw pseudo-labels generated by OWL-v2. 

To establish a baseline for comparison, we initially used OWL-v2 to perform inference on the entire image pool, i.e., Mapillary or nuImages datasets for each novel category. 
We measured the precision of the pseudo-labels for novel categories against the ground-truth labels in each dataset, considering a pseudo-label as a true positive if it achieved an Intersection over Union (IoU) greater than 0.5 with the ground truth. 
This baseline performance sets the stage for appreciating the enhancements brought by our \datafeeder{} and \modelupdater{}. Following this, we report on the precision of pseudo-labels after image-level filtering by our \datafeeder{} and pseudo-label filtering by our \modelupdater{}.

Table~\ref{tab:disect_datafeeder_model_updater} shows that compared to the raw pseudo-labels generated by OWL-v2, our \datafeeder{} alone improved the average precision of novel categories by 4.3\%. Furthermore, when combined with our \modelupdater{}, the average precision was enhanced to 45.7\%, which is a 24.3\% improvement over the raw pseudo-labels from OWL-v2. This significant improvement underscores the effectiveness of our AIDE in fine-tuning OWL-v2, surpassing the self-training method proposed by OWL-v2 in~\cite{minderer2023scaling}, as our AIDE provides substantially better quality pseudo-labels.

\begin{table}[t]
  \centering
    \footnotesize
    \begin{tabular}{cccc}
    \toprule
    {Category} & OWL-v2~\cite{minderer2023scaling} & \multicolumn{1}{l}{w/ \datafeeder{}} & \multicolumn{1}{l}{w/ \modelupdater{}} \\
    \midrule
    motorcyclist & 11.1  & 19.3  & 47.2 \\
    bicyclist & 5.3   & 7.6   & 33.8 \\
    const. vehicle & 11.3  & 12.8  & 16.5 \\
    trailer & 10.9  & 12.1  & 38.2 \\
    traffic cone & 68.3  & 76.9  & 92.9 \\
    \midrule
    Average AP(\%) & 21.4  & 25.7  & 45.7 \\
    \bottomrule
    \end{tabular}%
  \caption{Evaluate the quality of the pseudo-labels of novel categories generated by OWL-v2 without any post-processing, filtered by the \datafeeder{} with BLIP-2, and further filtered by \modelupdater{}. We measure the precision~(\%) by comparing the pseudo labels with ground-truth labels for each novel category. Given a pseudo-label, we treat it as a true positive if it has an IoU larger than 0.5 with the ground-truth label, otherwise it is a false positive.}
  \label{tab:disect_datafeeder_model_updater}%
\end{table}%

\section{Limitations}
 
Our work proposed the first automated data engine, AIDE, based on VLMs and LLMs for autonomous driving. 
However, there are still limitations in our work.
As AIDE is extensively integrated with VLMs and LLMs, the hallucination of VLMs and LLMs may have negative impacts on our \issuefinder{} and \verification{}. 
Although the dense captioning model in our \issuefinder{} can automatically identify the novel category with high precision, it may also potentially hallucinate novel categories that are not present in the image. On the other hand, although our \verification{} can generate diverse scene descriptions for evaluating our detector, it may also hallucinate scenarios that do not exist in the image pool. 

Generally, we believe that these concerns will be alleviated with the advancement of VLMs and LLMs in the future. 
Additionally, using a large image pool for text-based retrieval in \datafeeder{} can help mitigate these concerns. 
Despite the effectiveness of AIDE, for a safety-critical system, some human oversight is always recommended.

\section{More Experimental Details}
In this section, we provide more experimental details for our AIDE and also the comparison methods. For all approaches, including supervised training, semi-supervised learning, and AIDE, we begin with the same Faster RCNN model pretrained by the same six AV datasets then proceed to conduct our experiments. For the Unbiased Teacher-v1~\cite{liu2021unbiased}, we use the official implementation\footnote{\url{https://github.com/facebookresearch/unbiased-teacher}} and adhere to the same training settings. Both Supervised Training and AIDE are trained for 3000 iterations, using SGD optimization with a batch size of 4, a learning rate of 5e-4, and weight decay set at 1e-4 across all experiments. The Unbiased Teacher-v1~\cite{liu2021unbiased} requires a warm-up stage to pre-train a teacher model, so we allocate an additional 1000 iterations, totaling 4000 iterations, for training this method. All other training hyperparameters for the Unbiased Teacher-v1~\cite{liu2021unbiased} remain consistent with those used for Supervised Training and AIDE. For the image-text matching in \datafeeder{}, we leverage the `pretrain' configuration to initialize the BLIP-2 model, which is exactly based on the official BLIP-2 GitHub Repo\footnote{\url{https://github.com/salesforce/LAVIS/blob/main/examples/blip2_image_text_matching.ipynb}}. The VLMs we used are allowed for commercial usage (i.e., Otter/CLIP/BLIP-2). ChatGPT can be replaced by open-source LLMs like Llama2~\cite{touvron2023llama}, whereas the cost of ChatGPT is negligible~(less than \$0.01).

\subsection{Model Hyperparameters for \datafeeder{} and \modelupdater{}}

In this section, we detail the model hyperparameter selection for our \datafeeder{} and \modelupdater{}. Within our \datafeeder{}, we utilize BLIP-2 to query images relevant to each novel category. This is achieved by measuring the cosine similarity score between the text and image embeddings. Subsequently, all images are ranked based on their cosine similarity score (denoted as the BLIP-2 score), and the top-ranked images are selected by thresholding the BLIP-2 score. We have set the BLIP-2 score threshold at 0.6 for all novel categories. This threshold is chosen to ensure that our \datafeeder{} retrieves at least 1\% of the images from the image pool (comprising either Mapillary or nuImages datasets) for each novel category. Such a threshold guarantees that we have a sufficient number of images for pseudo-labeling in \modelupdater{}.

Second, in our \modelupdater{}, given that the number of relevant images has been significantly reduced following the BLIP-2 querying process (for example, only 550 images for ``motorcyclist''), we opt for a CLIP score threshold, specifically 0.1, for our two-stage pseudo-labeling to prevent excessive filtering out of too many potential pseudo-labels. As demonstrated in Section~\ref{dissect the impact} and Table~\ref{tab:disect_datafeeder_model_updater}, even with such a CLIP score threshold, we can still markedly enhance the quality of pseudo-labels compared to using only the \datafeeder{} to filter OWL-v2’s pseudo-labels. For filtering pseudo-labels of known categories, we set the confidence score threshold at 0.6. This threshold significantly reduces the number of pseudo-labels for each known category, helping to balance it with the number of pseudo-labels for novel categories. Such a balance is crucial in mitigating forgetting while simultaneously boosting performance for novel categories.

\subsection{Experimental Details for fine-tuning OWL-v2 with AIDE}
For the experiment of fine-tuning the OWL-v2~\cite{minderer2023scaling} with AIDE, we leverage the official model released by the author~\footnote{\url{https://github.com/google-research/scenic/tree/main/scenic/projects/owl_vit}}. We opted to use the Hugging Face Transformers library to fine-tune the OWL-v2~\footnote{\url{https://huggingface.co/docs/transformers/model_doc/owlv2}} as it provides a consistent codebase for both inferring and training OWL-v2 in PyTorch. Notably, the OWL-v2~\cite{minderer2023scaling} was self-training on the OWL-ViT~\cite{minderer2022simple} on a web-scale dataset, i.e., WebLI~\cite{chen2023pali}, and the fine-tuning learning rate is 2e-6. To enable effective continual fine-tuning with AIDE, we set the initial learning rate as 1e-7. This setting is intended to prevent dramatic changes in the weights of OWL-v2, thereby avoiding catastrophic forgetting while still allowing the model to learn novel categories using AIDE effectively. We utilize the same training hyperparameters from the self-training recipe of OWL-v2~\cite{minderer2023scaling} to conduct self-training of OWL-v2 on AV datasets in Section~\ref{sec: update AIDE}, ensuring a fair comparison.

\subsection{Details for our \verification{}}
\begin{figure}[t]
  \centering
  \includegraphics[width=0.45\textwidth]{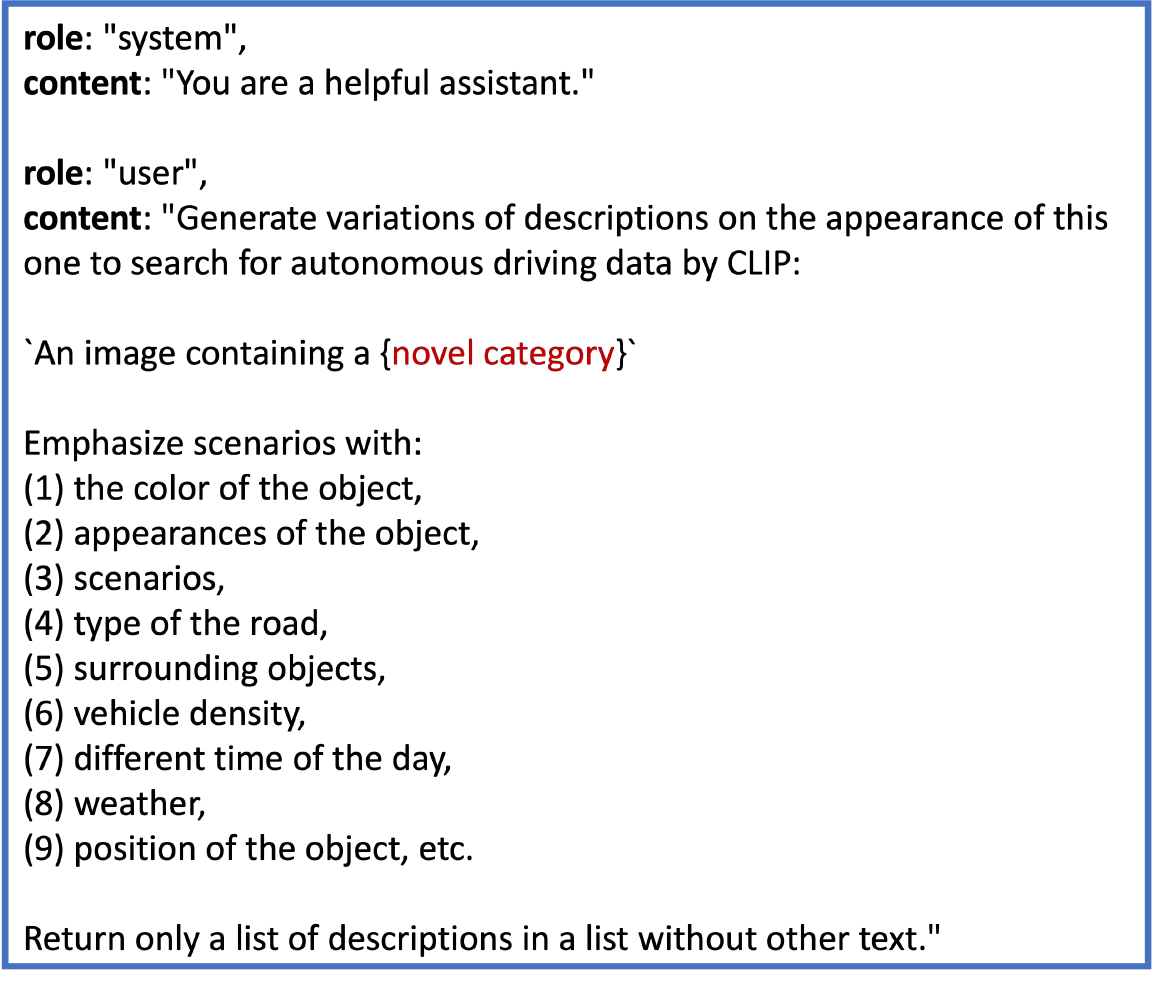}
  \caption{Prompt template for ChatGPT to generate diverse testing scenarios in \verification{}. The ``novel category'' is a placeholder in the template and will be replaced by the exact name of the novel category obtained in \issuefinder{}.}
  \label{fig: prompt template}
\end{figure}

As mentioned in our main paper Sec.~3.4, we leverage LLM, i.e., ChatGPT~\cite{ChatGPT}, to generate diverse scene descriptions to evaluate the updated detector from our \modelupdater{}. The prompt template we use for this purpose is illustrated in Figure~\ref{fig: prompt template}. Further, we have detailed the training process triggered by \verification{} in Section~\ref{sec: update AIDE}. We use the same training and model hyperparameters for our continual training in \modelupdater{} when conducting the training triggered by \verification{}. 

\section{More Visualizations}
\subsection{Predictions with Different Methods}

\begin{figure*}[h]
  \centering
  \includegraphics[width=0.85\textwidth]{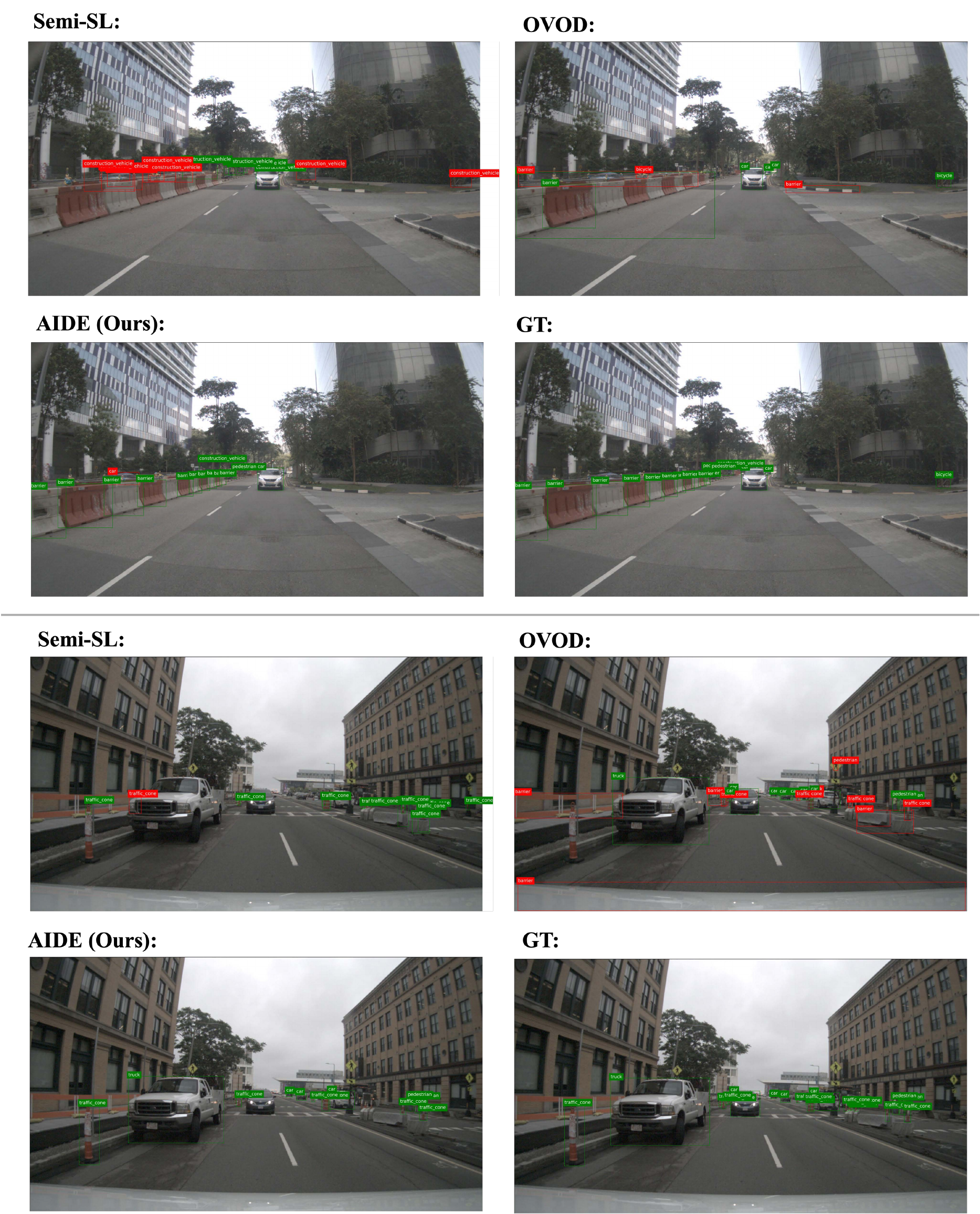}
  \caption{Visualization of the detection results under different methods. We treat a box prediction as true positive if it has an IoU larger than 0.5 with the ground-truth box. The true positive predictions are in {\color{Green}green} color, while the false positive predictions are in {\color{red}red} color. \textbf{Top-left}: Semi-supervised Learning~(Semi-SL) method, i.e., Unbiased Teacher-v1~\cite{liu2021unbiased}. \textbf{Top-right}: Open-vocabulary object detection~(OVOD) method, i.e., OWL-v2~\cite{minderer2023scaling}. \textbf{Bottom-left}: AIDE. Bottom-right: ground-truth.}
  \label{fig: combine 1}
\end{figure*}

\begin{figure*}[h]
  \centering
  \includegraphics[width=0.85\textwidth]{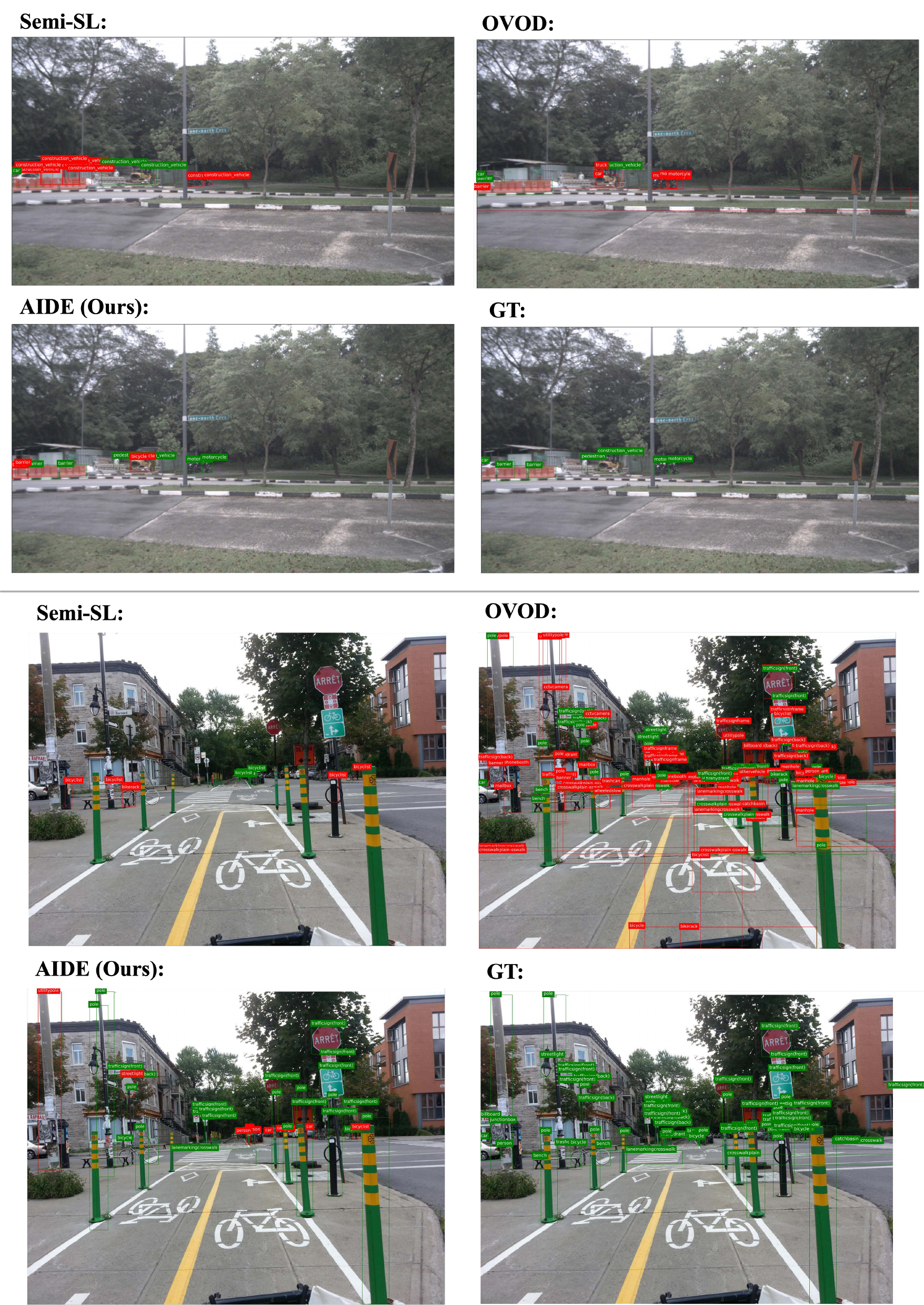}
  \caption{Visualization of the detection results under different methods. We treat a box prediction as true positive if it has an IoU larger than 0.5 with the ground-truth box. The true positive predictions are in {\color{Green}green} color, while the false positive predictions are in {\color{red}red} color. \textbf{Top-left}: Semi-supervised Learning~(Semi-SL) method, i.e., Unbiased Teacher-v1~\cite{liu2021unbiased}. \textbf{Top-right}: Open-vocabulary object detection~(OVOD) method, i.e., OWL-v2~\cite{minderer2023scaling}. \textbf{Bottom-left}: AIDE. \textbf{Bottom-right}: ground-truth.}
  \label{fig: combine 2}
\end{figure*}

\begin{figure*}[h]
  \centering
  \includegraphics[width=0.85\textwidth]{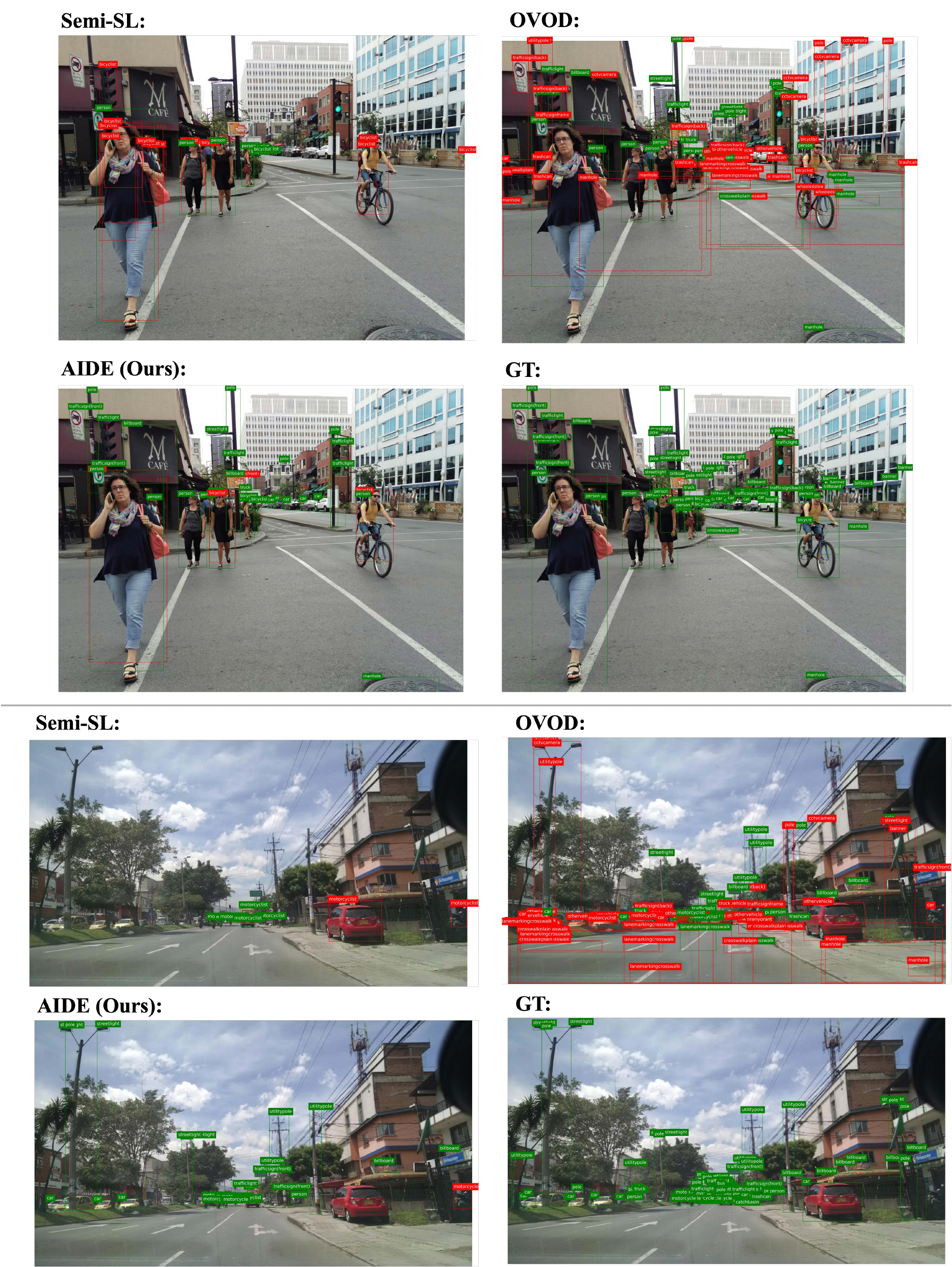}
  \caption{Visualization of the detection results under different methods. We treat a box prediction as true positive if it has an IoU larger than 0.5 with the ground-truth box. The true positive predictions are in {\color{Green}green} color, while the false positive predictions are in {\color{red}red} color. \textbf{Top-left}: Semi-supervised Learning~(Semi-SL) method, i.e., Unbiased Teacher-v1~\cite{liu2021unbiased}. \textbf{Top-right}: Open-vocabulary object detection~(OVOD) method, i.e., OWL-v2~\cite{minderer2023scaling}. \textbf{Bottom-left}: AIDE. \textbf{Bottom-right}: ground-truth. Note that some original annotations in Mapillary are not correct. For instance, for the image of ``GT'' in the second row, the human on the bicycle should be labeled as ``bicyclist'' while the original label is ``person''.}
  \label{fig: combine 3}
\end{figure*}

We present additional visualization results in Figures~\ref{fig: combine 1}, \ref{fig: combine 2}, and \ref{fig: combine 3}. These visualizations reveal that the Semi-Supervised Learning (Semi-SL) method tends to overfit to novel categories, resulting in numerous false positive predictions. Furthermore, the Semi-SL method struggles to detect known categories, indicating an issue with catastrophic forgetting. In contrast, the state-of-the-art Open-Vocabulary Object Detection (OVOD) method, specifically OWL-v2, also produces many false positives for both novel and known categories. However, compared to both the Semi-SL and OVOD methods, AIDE demonstrates superior performance in accurately detecting both novel and known categories.

\subsection{Prediction after updating our model by \verification{}}

\begin{figure*}[h]
  \centering
  \includegraphics[width=\textwidth]{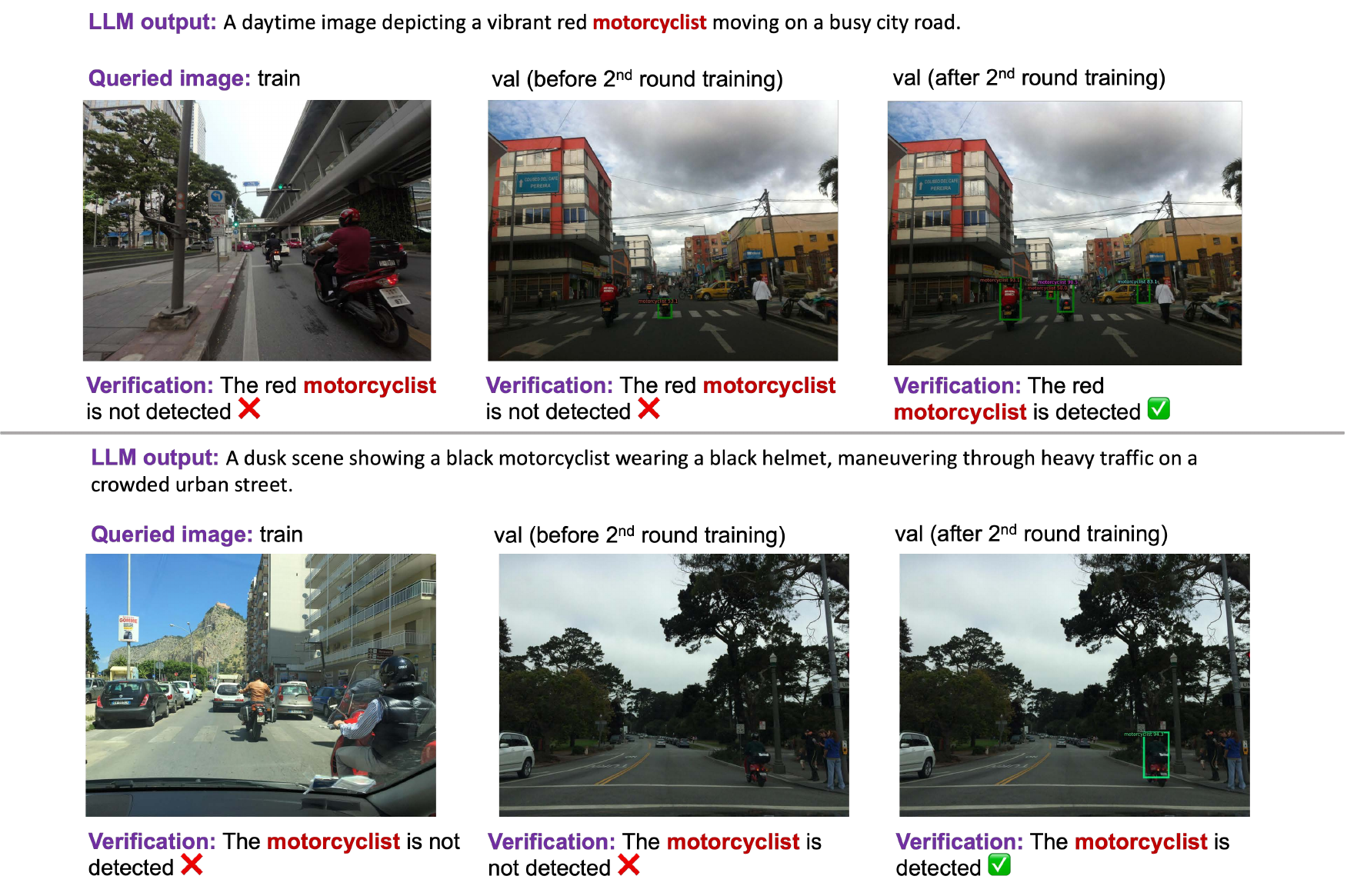}
  \caption{More visualizations on our \verification{}. \textbf{Left:} In the queried image from the training set for verification, the model is not predicting the motorcyclist.
  \textbf{Middle:} Similarly on the queried image from the validation set, the model is not predicting the motorcyclist.
  \textbf{Right:} After updating the model again, our model can successfully predict the motorcyclist. }
  \label{fig: verf update}
\end{figure*}

In Figure~\ref{fig: verf update}, we present additional visualizations to Fig.~7 in our main paper to demonstrate that an extra round of training, initiated by \verification{}, further reduces both missed and incorrect detections of novel categories. These visualizations illustrate the effectiveness of the additional training round in enhancing the accuracy and reliability of our detection system for these novel categories.

\section{Discuss about de-duplication process for video data}
The nuImages dataset contains 13 frames per scene, spaced 0.5 seconds apart. Currently, we directly use all unlabeled images of nuImages dataset for \datafeeder{} to query without using any de-duplication process in our main paper. In practice, as the dataset gets larger or with a higher frame rate, de-duplication could further improve the data diversity for querying in \datafeeder{} and may potentially improve the performance of AIDE, and we leave this for future study.

\section{Comparison between \verification{} and Active Learning alternatives}
We compare our approach, ``LLM description+BLIP-2'' for \verification{}, with two Active Learning~(AL) baselines. The first one is to verify the boxes predicted as the novel target class by the detector but with the highest classification entropy. The second one is to perform verification on randomly sampled boxes predicted as the novel target class by the detector. For both AL baselines, we use them to verify 10 images, the same as what we have done in Sec.~4.3.4 of our main paper. The two AL baselines only achieve 13.1\% and 12.7\% AP on novel classes, respectively. This is inferior to our approach (14.2\% AP) which uses VLM/LLM to identify diverse AV scenarios for verification.

\section{Discussion for the real-cost of supervised and semi-supervised methods}
In our main paper Fig.~1, Tab.~1, and Tab.~2, we only measure the ``Labeling and Training'' cost for the supervised/semi-supervised methods. In fact, the real cost for the supervised/semi-supervised method is not just labeling images but also includes \emph{searching over the large data pool to find relevant images} to label. For instance, an annotator needs to examine 874 images on average to find 50 images for a selected novel class, costing \$43.7 for supervised/semi-supervised methods, assuming it costs 10 seconds per image to inspect for novel classes, which corresponds to \$0.05 at \$18 per hour. Therefore, AIDE is more practical than supervised/semi-supervised methods for car companies as we automate data querying in \datafeeder{} to largely reduce the total cost.

\newpage

\begin{table*}[h]
\centering
\footnotesize
\begin{tabular}{c|c|c|c|c|c|c}
\toprule
 & Cityscapes & KITTI & BDD100k & nuImages & Mapillary & Waymo \\
\midrule
\#~Classes & 8 & 3 & 10 & 10 & 37 & 3 \\
Cumulative \#~Classes & 8 & 10 & 12 & 16 & 45 & 46 \\
\#~Images & 2,975 & 6,859 & 69,863 & 67,279 & 18,000 & 790,405 \\
\midrule
\multirow{12}{*}{Vehicle} & car & car & car & car & car & \\
 & truck & & truck & truck & truck & \\
 & bus & & bus & bus & bus & \\
 & train & & train & & & \\
 & motorcycle & & motorcycle & motorcycle & motorcycle & \\
 & bicycle & & bicycle & bicycle & bicycle & \\
 & & & & construction vehicle & & \\
 & & & & trailer & trailer & \\
 & & & & & caravan & \\
 & & & & & boat  &\\
 & & & & & wheeled-slow & \\
 & & & & & other vehicle & \\
 & & & & & & vehicle \\
\hline
\multirow{5}{*}{Human} & person & & & & person & \\
 & & pedestrian & pedestrian & pedestrian & & pedestrian \\
 & rider & & rider & motorcyclist & & \\
 & & cyclist & & & bicyclist & cyclist \\
 & & & & & other rider & \\
\hline
\multirow{9}{*}{Traffic Objects} & & & & traffic cone &  & \\
 & & & & barrier &  & \\
 & & & traffic light & & traffic light &  \\
 & & & traffic sign & &  traffic sign(back) & \\
 & & & & & traffic sign(front) & \\
 & & & & traffic sign frame  &\\
 & & & & & pole  & \\
 & & & & & street light & \\
 & & & & & utility pole & \\
\hline
\multirow{16}{*}{Other Objects} & & & & & bird & \\
 & & & & & ground animal & \\
 & & & & & crosswalk plain & \\
 & & & & & lane marking crosswalk &  \\
 & & & & & banner & \\
 & & & & & bench &\\
 & & & & & bike rack & \\
 & & & & & billboard & \\
 & & & & & catch basin & \\
 & & & & & cctv camera &  \\
 & & & & & fire hydrant & \\
 & & & & & junction box & \\
 & & & & & mailbox & \\
 & & & & & manhole & \\
 & & & & & phone booth & \\
 & & & & & trash can & \\
\bottomrule
\end{tabular}
\caption{The statistics and label space of the six AV datasets, i.e., Cityscapes~\cite{Cordts2016Cityscapes}, KITTI~\cite{Geiger2012CVPR}, BDD100k~\cite{bdd100k}, nuImages~\cite{nuscenes2019}, Mapillary~\cite{neuhold2017mapillary}, and Waymo~\cite{Sun_2020_CVPR}. There are 46 categories in total after combining the label spaces. 
To simulate the novel categories and ensure that the selected categories are meaningful and crucial for AV in the street, we choose 5 categories as novel categories: ``motorcyclist'' and ``bicyclist'' from Mapillary, ``construction vehicle'', ``trailer'', and ``traffic cone'' from nuImages.
The rest 41 categories are set as known. We remove all the annotations for these categories in our joint datasets and also remove the related categories with similar semantic meanings, e.g., ``bicyclist'' vs ``cyclist'', ``rider'' vs ``motorcyclist''.}
\label{tab:datasets_labelspace}
\end{table*}

\end{document}